\documentclass{article}
\usepackage{amssymb,amsmath,amsfonts,bm}
\usepackage[final]{corl_2025} 

\usepackage{graphicx}
\usepackage[table]{xcolor}  
\usepackage{hyperref}       
\usepackage{cleveref}       
\definecolor{darkblue}{rgb}{0.0, 0.0, 0.5}
\hypersetup{
  colorlinks=true,
  linkcolor=darkblue,
  citecolor=darkblue,
  urlcolor=darkblue
}
\usepackage{mathtools}
\usepackage{wrapfig}
\usepackage{algorithmic}
\usepackage{algorithm}

\usepackage{amsthm}
\usepackage{booktabs}
\usepackage{adjustbox}
\newtheorem{definition}{Definition}[section]
\newtheorem{note}{Note}[section]
\newtheorem{proposition}{Proposition}[section]
\theoremstyle{plain}

\crefname{figure}{Fig.}{Figs.}
\crefname{equation}{Eq.}{Eqs.}
\crefname{definition}{Def.}{Defs.}
\crefname{proposition}{Prop.}{Props.}
\crefname{table}{Tab.}{Tabs.}
\crefname{section}{Sec.}{Secs.}
\crefname{appendix}{Appx.}{Appxs.}

\def\1{\bm{1}}

\def\vaa{{\bm{a}}}

\def\vo{{\bm{o}}}

\def\vr{{\bm{r}}}
\def\vs{{\bm{s}}}

\def\vx{{\bm{x}}}
\def\vy{{\bm{y}}}

\def\vsA{{\mathcal{A}}}

\def\vsO{{\mathcal{O}}}

\def\vsS{{\mathcal{S}}}

\def\vsX{{\mathcal{X}}}
\def\vsY{{\mathcal{Y}}}

\def\mI{{\bm{I}}}

\DeclareMathAlphabet{\mathsfit}{\encodingdefault}{\sfdefault}{m}{sl}
\SetMathAlphabet{\mathsfit}{bold}{\encodingdefault}{\sfdefault}{bx}{n}

\def\sD{{\mathbb{D}}}

\def\sI{{\mathbb{I}}}

\def\sK{{\mathbb{K}}}

\def\sN{{\mathbb{N}}}

\newcommand{\R}{\mathbb{R}}

\usepackage{relsize}
\usepackage{exscale}

\newcommand{\G}[1][]{\mathbb{G}_{\scalebox{0.6}{$#1$}}}       
\newcommand{\g}{g}       
\newcommand{\Glact}[1][]{
	\!\,
    \mathrel{
		\mathsmaller{
			\triangleright_{\scalebox{0.5}{$#1$}}
		}
	}
    \!\,
}             
\newcommand{\Gract}[1][]{\mathrel{\mathsmaller{\triangleleft}}}              
\newcommand{\Gconj}[1][]{\!\,\mathrel{\mathsmaller{\diamond}}}                   
\newcommand{\Gcomp}[1][]{\!\,\mathrel{\mathsmaller{\circ}}\!\,}                      

\newcommand{\CyclicGroup}[1][]{\mathbb{C}_{#1}}        

\newcommand{\GLGroup}{\mathbb{GL}}          
\newcommand{\SO}[1][3]{\mathbb{SO}_{#1}}                 
\newcommand{\SE}[1][3]{\mathbb{SE}_{#1}}                 
\newcommand{\EG}[1][3]{\mathbb{E}_{#1}}

\newcommand{\rep}[1][]{
\bm{\rho}_{
\scalebox{0.65}{$#1$}
}
}

\newcommand{\bSet}[1]{\sI_{\scalebox{0.8}{$#1$}}}  

\newcommand{\mapping}[5]{ 
	\begin{matrix}
		#1: & #2 & \longrightarrow & #3 \\
		    & #4 & \longrightarrow & #5
	\end{matrix}
}

\usepackage[all]{xy}

\newcommand{\homomorphismDiag}[5]{
	\xymatrix{
		#1 \ar@{-}[r]^{#3}    \ar[d]^{#5} & #1 \ar[d]^{#5} \\
		#2 \ar@{-}[r]^{#4}                   & #2
	}
}
\newcommand{\invariantDiag}[5]{
	\xymatrix{
		#1 \ar@{-}[r]^{#3} \ar[dr]_{#4} & #1 \ar[d]^{#4} \\
		& #2 \ar@(r,d)[]^{#5}
	}
}

\newcommand{\isomorphismDiag}[5]{
\xymatrix{
#1 \ar@{-}[r]^{#3}    \ar@<0.5ex>[d]^{#5} \ar@{<-}[d]_{#5^{-1}} & #1 \ar@<0.5ex>[d]^{#5} \ar@{<-}[d]_{#5^{-1}} \\
#2 \ar@{-}[r]^{#4}                   & #2
}
}

\newcommand{\mdpKernel}{\tau}
\newcommand{\obsSpace}{\vsO}
\newcommand{\obsFn}{\bm{\sigma}}
\newcommand{\initDist}{\rho_0}
\newcommand{\reward}{r}

\definecolor{gray}{rgb}{0.6, 0.7, 0.7}
\definecolor{awesomeblue}{rgb}{0.054, 0.415, 0.505}
\definecolor{awesomeorange}{rgb}{0.570, 0.458, 0.0912}

\newcommand{\ubcolor}[3][awesomeblue]{{
        \color{#1}{
            \underbrace{\color{black}{#2}}_{#3}
        }
    }}

\newcommand{\nnParams}{\bm{\theta}}
\newcommand{\nnParamsDist}{\bm{\phi}}

\newcommand{\chgAgent}[1]{\textcolor{blue}{#1}}
\newcommand{\chgTask}[1]{\textcolor{red}{#1}}
\newcommand{\rightN}{\textnormal{R}}
\newcommand{\leftN}{\textnormal{L}}
\newcommand{\bowl}{\textnormal{B}}
\newcommand{\egg}{\textnormal{E}}

\title{Morphologically Symmetric Reinforcement Learning for Ambidextrous Bimanual Manipulation
}

\author{
  Zechu Li~$^1$ ~~~~Yufeng Jin~$^{1,2}$ ~~~~Daniel Ordoñez Apraez~$^3$ ~~~~\textbf{Claudio Semini}~$^3$ \\
   \textbf{Puze Liu}~$^4$\thanks{Corresponding author} 
  ~~~~ \textbf{Georgia Chalvatzaki}~$^{1,5,6}$\\
  TU Darmstadt~$^1$~~~~Honda Research Institute Europe~$^2$
  ~~~~Istituto Italiano di Tecnologia~$^3$
  \\
  DFKI~$^4$
  ~~~~Hessian.AI~$^5$~~~~Robotics Institute Germany~$^6$
}
\usepackage[nonumberlist]{glossaries}  
\newacronym[
  shortplural = MDPs,                
  longplural = Markov Decision Processes,
  first={Markov Decision Process},    
  firstplural={Markov Decision Processes} 
]{mdp}{MDP}{Markov decision process}
\newacronym{nn}{NN}{Neural Network}
\newacronym{pomdp}{POMDP}{Partially Observable MDP}
\newacronym{mtmapomdp}{MTMA-POMDP}{Multi-Task Multi-Agent POMDP}
\newacronym{smmtmapomdp}{Morphologically Symmetric MTMA-POMDP}{Morphologically Symmetric Multi-Task Multi-Agent Partially Observable MDP}

\newacronym[
  shortplural = DoF,                
  longplural  = degrees of freedom,  
  first={degree of freedom (DoF)},   
  firstplural={degrees of freedom (DoF)}, 
]{dof}{DoF}{degree of freedom}
\newacronym{ik}{IK}{inverse kinematics}
\newacronym{rl}{RL}{Reinforcement Learning}
\newacronym{PPO}{PPO}{Proximal Policy Optimization}
\newacronym[
    name={PPO},
	description={
		Proximal Policy Optimization: desc if needed}
]{ppo}{PPO}{Proximal Policy Optimization}

\newacronym[
    name={ E-PPO },
    description={
		Equivariant Proximal Policy Optimization: A single policy with an equivariant network controlling the entire system jointly (44-DoF). Represents standard PPO with symmetry handling}
]{eppo}{E-PPO}{Equivariant PPO}

\newacronym[
    name={ IPPO },
    description={
		Independent Proximal Policy Optimization: Learns two fixed policies, each assigned to one arm. Due to scene randomization, each policy must learn both subtasks and select the appropriate one}
]{ippo}{IPPO}{Independent IPPO}

\newacronym[
    name={ E-IPPO },
    description={
		Equivariant Independent Proximal Policy Optimization: Trains a single policy for one arm (22-DoF) and applies it to both arms. The symmetry effectively doubles the training data and includes task encoding in the observation to facilitate learning}
]{eippo}{E-IPPO}{Equivariant IPPO}

\newacronym[
    name={ SM-c },
    description={
		SYMDEX-c: Matches our method's policy structure but uses a global value function instead of separate ones, ablating the effect of global value functions commonly used in multi-agent cooperative settings}
]{smc}{SM-c}{SYMDEX-c}

\newacronym[
    name={ SM-aug },
    description={
		SYMDEX-aug: Uses the same training scheme as our method but replaces the equivariant network with on-policy data augmentation using group transformations, a common alternative in symmetry learning}
]{smaug}{SM-aug}{SYMDEX-aug}

\makeglossaries

\usepackage{float}

\begin{document}
\maketitle

\begin{figure}[H]
	\centering
	\vspace{-1.0em}
	\includegraphics[width=0.9\textwidth]{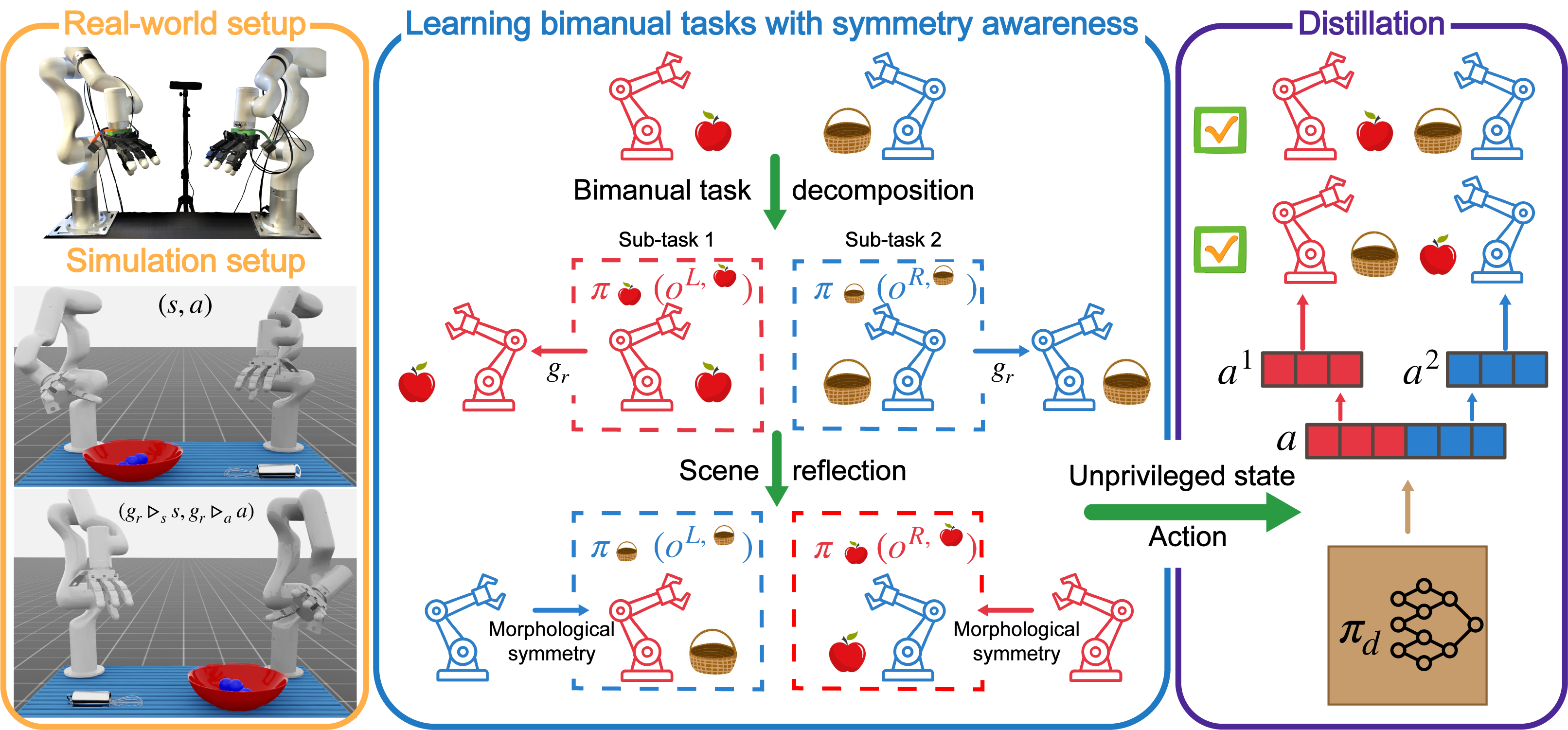}
	\caption{Overview of SYMDEX: (Left) Digital twin of our bimanual robot. (Middle) The task is decomposed into two sub-tasks, each trained with a dedicated equivariant policy that  transfers across symmetric configurations. (Right) Task-specific policies are distilled into an equivariant policy.}
	\label{fig:overview}
\end{figure}

\begin{abstract}
	Humans naturally exhibit bilateral symmetry in their gross manipulation skills, effortlessly mirroring simple actions between left and right hands. Bimanual robots—which also feature bilateral symmetry—should similarly exploit this property to perform tasks with either hand. Unlike humans, who often favor a dominant hand for fine dexterous skills, robots should ideally execute ambidextrous manipulation with equal proficiency. To this end, we introduce \textsc{SYMDEX} (SYMmetric DEXterity), a reinforcement learning framework for ambidextrous bi-manipulation that leverages the robot's inherent bilateral symmetry as an inductive bias. \textsc{SYMDEX} decomposes complex bimanual manipulation tasks into per-hand subtasks and trains dedicated policies for each. By exploiting bilateral symmetry via equivariant neural networks, experience from one arm is inherently leveraged by the opposite arm. We then distill the subtask policies into a global ambidextrous policy that is independent of the hand-task assignment. We evaluate \textsc{SYMDEX} on six challenging simulated manipulation tasks and demonstrate successful real-world deployment on two of them. Our approach strongly outperforms baselines on complex task in which the left and right hands perform different roles. We further demonstrate SYMDEX’s scalability by extending it to a four-arm manipulation setup, where our symmetry-aware policies enable effective multi-arm collaboration and coordination. Our website is made publicly available at: \url{https://supersglzc.github.io/projects/symdex/}.    
\end{abstract}


\section{Introduction}
\vspace{-0.8em}
Humans inherently exhibit \href{https://bit.ly/bilateral-symmetry}{bilateral symmetry} in their gross motor skills, which allows them to effortlessly mirror simple actions between their left and right limbs. However, when it comes to fine dexterous tasks (e.g., writing, playing instruments), most people develop a dominant side, a phenomenon known as \textit{handedness}. This functional control asymmetry often leads to suboptimal task strategies, such as switching hands to maintain control robustness. In contrast, bimanual robots---which frequently also feature bilateral symmetry---are not inherently bound by handedness. Hence, in the context of manipulation, there is a unique opportunity to design algorithms that perform tasks \href{https://en.wikipedia.org/wiki/Ambidexterity}{ambidextrously}, which enables the interchangeable use of limbs across a wide range of task configurations and plans actions based on efficiency rather than a left/right preference.

Achieving ambidextrous bimanual manipulation requires control policies incorporating awareness of the robot's bilateral symmetry. Recent robotics research has explored such structural symmetries—referred to as \textit{morphological symmetries} \cite{ordonez2025morphological}—to develop symmetry-aware learning methods. Most previous work has focused on legged locomotion, where exploiting morphological symmetry improves control robustness and sample efficiency \cite{su2024leveraging,ordonez2023rss,mittal2024symmetry,bao2025toward,butterfield2024mi,xie2024morphological,zhao2025hierarchical}. In contrast, its use in manipulation—especially for bimanual or multi-robot systems—remains uncharted \cite{amadio2019exploiting}. Whether embedding symmetry in manipulation policies can offer similar gains in generalization and sample efficiency for high-dimensional, contact-rich tasks is still an open question.

\gls{rl} is a compelling paradigm for bimanual dexterous manipulation, especially in sim-to-real settings \cite{lin2024twisting,huang2023dynamic,lin2025sim}. Unlike imitation learning, which needs large, high-quality demonstrations, \gls{rl} trains in randomized environments, acquiring robust behaviors via massive parallel simulation. Yet the complexity of bimanual or multi-robot systems has confined prior work to narrowly scoped tasks enforced by system constraints (e.g., hand-only control \citep{lin2024twisting} or arm joint locking \citep{huang2023dynamic}). Hence we ask: \textit{Can \gls{rl} scale to fully actuated bimanual—and multi-robot—systems by embedding morphological symmetry as a structural prior in policy learning?}

Learning bimanual (and multi-arm) manipulation via \gls{rl} presents significant challenges. First, the \textbf{high-dimensional observation and action spaces} make policy learning difficult: a bimanual robot with dexterous hands must jointly process and control numerous joints, leading to an exponentially large exploration space~\citep{lin2024twisting,huang2023dynamic,lin2025sim,lan2023dexcatch}. Second, the dual-arm setup and task complexity exacerbate the \textbf{credit assignment problem}~\citep{jiang2025rethinking,wang2022individual}. During exploration, one arm may succeed while others fail, and the reward signal merges both positive and negative feedback, making reward shaping intractable in the presence of numerous task-specific rewards. Additionally, when ambidexterity is involved, each arm may perform different tasks, turning the problem into \textbf{multi-task learning}. Although individual reward terms can be well-defined, the scale of each term needs careful tuning to balance learning across tasks. Finally, \textbf{zero-shot sim-to-real transfer} introduces challenges such as the sim-to-real gap and safety concerns, requiring the learned policy to at least prevent inter-arm collisions during deployment~\citep{chen2023visual, tiboni2023domain,akkaya2019solving}. We provide a detailed review of related work in Appx.~\ref{sec:related_work}.

To address these open challenges, we introduce \textbf{SYMDEX} (SYMmetric DEXterity), a \gls{rl} framework for ambidextrous bimanual (and multi-arm) dexterous manipulation, that explicitly incorporates morphological symmetry as an inductive bias, both architecturally and algorithmically. \textsc{SYMDEX} decomposes complex bimanual tasks into per-hand subtasks and trains a separate policy for each using an equivariant neural network \citep{bronstein2021geometric}. This structure inherently shares experience across symmetric limbs, exploiting morphological symmetry to accelerate learning. \textsc{SYMDEX} operates entirely in joint space, without relying on task-space solvers or handcrafted action symmetries. To enable flexibility and remove the need for fixed hand-task assignment, we distill these sub-policies into a unified global equivariant policy via teacher-student distillation. The resulting policy is ambidextrous by design and zero-shot deployable in the real world.

Our contributions are as follows: \textbf{1. Morphological symmetry-aware policy learning approach.} We present \textsc{SYMDEX}, a \gls{rl} framework that explicitly leverages the inherent morphological symmetry in bimanual robotic systems to enable ambidextrous control and generalization across structurally symmetric task configurations.
\textbf{2. Scalable framework generalizable to multi-arm tasks.} We demonstrate that our approach can be deployed to more complex multi-arm settings with complexer symmetry groups without increasing the task complexity. \textbf{3. A full learning recipe enables zero-shot sim-to-real transfer}. Our framework addresses key challenges to ensure robust real-world deployment, we incorporate a curriculum learning strategy for zero-shot sim-to-real transfer. We evaluate \textsc{SYMDEX} on six diverse and challenging bimanual manipulation tasks in simulation and successfully deploy it on two of them in the real world.
\vspace{-1em}
\section{Background}
\label{sec:background}
\vspace{-0.5em}
\noindent
Here, we review the foundational concepts and notation necessary for formalizing how symmetries serve as an inductive bias in learning bimanual (and multi-robot) dexterous manipulation policies via \gls{rl}. Extended definitions are provided in Appx.~\ref{sec:group_theory}.

A \textbf{symmetry group} (see \cref{def:group}) is a set of invertible transformations, denoted as $\G = \{e, g_a, g_b, \dots\}$, that can be defined to act on distinct objects, such as the state $\vsS$ and action $\vsA$ spaces of a \gls{mdp}. To do so we define the \textbf{group actions} (see \cref{def:left_group_action}). Specifically, let
$ (\Glact[\vsS]) \colon \G \times \vsS \to \vsS $
and
$ (\Glact[\vsA]) \colon \G \times \vsA \to \vsA $
denote the action of $\G$ on $\vsS$ and $\vsA$, respectively. Then, given a symmetry transformation $\g \in \G$ and a state-action pair $(s, a) \in \vsS \times \vsA$, the $g$-transformed pair is denoted by $(\g \Glact[\vsS] s, \g \Glact[\vsA] a) \in \vsS \times \vsA$ (see \cref{fig:overview}-left).




The \textbf{symmetries of \glspl{mdp}} are defined as state–action transformations that preserve the \gls{mdp}'s dynamics. This property is characterized by the $\G$-equivariance (see \cref{def:equivariantMaps}) of the dynamics:
\begin{equation}
	\label{eq:mdp_dyn_equivariance}
	\small
	\g\Glact[\vsS]\mathbb{E}[f(s,a)] = \mathbb{E}[f(\g\Glact[\vsS]s,\;\g\Glact[\vsA]a)],
	\qquad\forall(s,a)\in\vsS\times\vsA,\;\forall\g\in\G,
\end{equation}
where $f\!:\! \vsS \times \vsA \to \vsS$ is a function mapping state–action pairs to subsequent states. For example, consider the bimanual environment in \cref{fig:overview}, where the symmetry group is the reflection group $\G = \CyclicGroup[2] = \{e, g_r \mid g_r^2 \!=\! e\}$—with $g_r$ denoting the robots' bilateral symmetry. Here, \cref{eq:mdp_dyn_equivariance} shows that the in-hand manipulation dynamics for the left and right hands are equivalent, up to $g_r$.

The symmetry priors from \cref{eq:mdp_dyn_equivariance} constrain the \gls{mdp}'s optimal policy and value function. To see this, let's formally denote a \gls{pomdp} by the tuple
$\langle \vsS, \vsA, r, \mdpKernel, \initDist, \gamma, \obsSpace, \obsFn \rangle$\footnote{
	In manipulation tasks, observations of number of contacts and contact points are typically unavailable, hence the control problem of manipulation is modeled as a \gls{pomdp}.
},
where $\vsS$, $\vsA$, and $\obsSpace$ are the state, action, and observation spaces;
$r: \vsS \times \vsA \to \R$ is the reward function;
$\mdpKernel: \vsS \times \vsA \times \vsS \to \R_+$ is the transition kernel;
$\initDist: \vsS \to \R_+$ is the initial state distribution;
$\gamma$ is the discount factor; and
$\obsFn: \vsS \to \obsSpace$ is the observation function.
A \gls{pomdp} is said to be \textbf{symmetric} if the following conditions hold:
\begin{definition}[Symmetric \gls{pomdp}]
	\label{def:symm_mdp}
	A \gls{pomdp} $\langle \vsS, \vsA, r, \mdpKernel, \initDist, \gamma, \obsSpace, \obsFn \rangle$ possess the symmetry group $\G$ when the state and action spaces $\vsS$ and $\vsA$ admit group actions $(\Glact[\vsS])$ and $(\Glact[\vsA])$, and $(r, \mdpKernel, \initDist)$ are all $\G$-invariant. That is, if for every $\g\in\G$, $\vs,\vs'\in\vsS$, and $\vaa\in\vsA$, we have:
	\begin{equation}
		\label{eq:symm_mdp}
		\small
		\mdpKernel(\g \Glact[\vsS] \vs' \mid \g \Glact[\vsS] \vs, \g \Glact[\vsA] \vaa) = \mdpKernel(\vs' \mid \vs, \vaa),
		\quad\;\;\;
		\initDist(\g \Glact[\vsS] \vs) = \initDist(\vs),
		\quad\;\;
		r(\g \Glact[\vsS] \vs, \g \Glact[\vsA] \vaa) = r(\vs, \vaa).
	\end{equation}
	\gls{pomdp}'s satisfying \cref{eq:symm_mdp} are constrained to have \textbf{optimal} policy and value functions satisfying:
	\begin{equation}
		\label{eq:symm_mdp_sol}
		\small
		\ubcolor{
			\g \Glact[\vsA] \pi^*(\obsFn(s)) = \pi^*(\obsFn(\g \Glact[\vsS] s))}
		{\text{\tiny Policy $\G$-equivariance}},
		\qquad
		\ubcolor[awesomeorange]{
			V^{*}(\obsFn(s)) = V^{*}(\obsFn(\g \Glact[\vsS] s))
		}{\text{\tiny Value function $\G$-invariance}}
		,
		\quad
		\forall\; s \in \vsS,\;\g \in \G. \;
		\text{(refer to \citep{zinkevich2001symmetry_mdp_implications})}
	\end{equation}
	\begin{note}
		\label{note:conditions_for_optimality}
		A policy $\pi$ of a $\G$-symmetric \gls{pomdp} can satisfy the $\G$-equivariance constraints of \cref{eq:symm_mdp_sol} only if the observation function $\obsFn$ is $\G$-equivariant (see \cref{prop:conditions_for_optimality_appx}). Therefore, to leverage \cref{eq:symm_mdp_sol} in policy learning, $\obsSpace$ must carry a group action $(\Glact[\obsSpace])$ and $\obsFn$ must be $\G$-equivariant.
	\end{note}
\end{definition}

\paragraph{Bimanual and multi-robot manipulation}
In bimanual (and multi-robot) dexterous manipulation, each task (e.g., stir eggs; see \cref{fig:overview}) can be decomposed into a sequence of concurrent and sequential subtasks, with each agent assigned subtasks (e.g., left arm grasps the egg beater while right arm holds the bowl). Hence, these environments are modeled as a \gls{mtmapomdp} 
defined by the tuple
$
	\langle \vsS, \vsA, R, \mdpKernel, \initDist, \gamma, \obsSpace, \obsFn, \sK, \sN \rangle
$,
where $\sN$ denotes the agent set—with $n\in\sN$ representing a unique robot arm (with a dexterous hand)—and $\sK$ denotes the task set—with $k\in\sK$ a manipulation subtask. This structure enables decomposition of the overall action space as $\vsA = \oplus_{n \in \sN} \vsA_{n}$, and defines subtask policies
$\vaa^{n} \sim \pi_{k}(\vo^{n,k}) \in \vsA_{n}$ for all $k\in\sK$, where $\vo^{n,k} = \obsFn^{n}(\vs, k)$ denotes the subtask-and-agent specific observation. Each task defines a reward $r_{k}$, which define the corresponding value function
$V^{k}(\vo_t^{n,k}) = \mathbb{E}_{\pi_{k}}\left[\sum_t^\infty \gamma^t r_{k}(\vo_t^{n,k})\right]$.
Consequently, the \gls{mtmapomdp} reward and value functions are defined as:
$
	\reward(\vs_t) \!=\! \sum_{(n,k) \in \sI} \reward_{k}(\obsFn^{n}(\vs_t, k))
$
and
$
	V(\vs_t) \!=\! \sum_{(n,k) \in \sI} V_{k}(\obsFn^{n}(\vs_t, k))
$.
Where $\sI$ denotes the set of agent-subtask pairwise pairings.

\vspace{-0.5em}
\section{Method}
\vspace{-0.5em}
\label{sec:method_intro}
In this paper, we aim to achieve ambidextrous bimanual manipulation—enabling a robot to perform tasks equally well with either arm and choose actions based on efficiency. This requires each arm to flexibly handle different subtasks depending on scene configuration. For example, in a bimanual scenario (see \cref{fig:overview}-left-middle), the robot uses its right arm to operate an egg beater while its left holds a bowl; in a reflected workspace (see \cref{fig:overview}-left-bottom), the bowl is closer to the right arm and the egg beater to the left. Consequently, the optimal behavior is to switch roles—using the right arm to hold the bowl and the left to operate the egg beater.

Learning such ambidextrous policy is challenging due to the high-dimensional observation-action spaces and the difficulty of credit assignment between arms. To address this, we formulate bimanual manipulation as a \gls{mtmapomdp} (Sec.~\ref{sec:background}), where each agent corresponds to a single robot arm executing one subtask.
This reduces the dimensionality of each agent's observation-action spaces and assigns subtask-specific reward, simplifying credit assignment. However, each agent must still learn to perform all subtasks to achieve ambidexterity. Notably, there is symmetry between the subtasks assigned to each agent (\cref{fig:overview}), which motivates leveraging morphological symmetries as a strong inductive bias and learning an equivariant policy for each subtask.

\paragraph{An illustrative example}
To express this ambidexterity using the formalism of \cref{sec:background}, note that changes in  agents' subtask assignments are formalized through group action on set of agent-task pairs $\sI$  (see \cref{def:left_group_action}), i.e.,
$
	(\Glact[\sI]) \colon \G \times (\sN \times \sK) \to (\sN \times \sK).
$
Thus, in the bimanual manipulation environment of \cref{fig:overview}, with $\sN\!=\!\{\rightN,\leftN\}$ and $\sK\!=\!\{\bowl, \egg\}$—where $\rightN$ and $\leftN$ denote the left/right arms, and $\bowl$ and $\egg$ denote the bowl-holding and egg-beater-operating subtasks—a bilateral reflection of the workspace, $g_r$, leads to the following permutation of \chgAgent{agents} and \chgTask{tasks}:
\begin{equation}
	\small
	\begin{aligned}
		g_r \!\Glact[\sI ] \!
		\begin{bmatrix}
			(\chgAgent{\leftN}, \chgTask{\bowl}) \\
			(\chgAgent{\rightN}, \chgTask{\egg})
		\end{bmatrix}
		\!=\!
            \begin{bmatrix}
			(\chgAgent{\leftN}, \chgTask{g_r \!\Glact[\sK ] \! \bowl}) \\
			(\chgAgent{\rightN}, \chgTask{g_r \!\Glact[\sK ] \!\egg})
		\end{bmatrix}
            \!=\!
            \begin{bmatrix}
			(\chgAgent{\leftN}, \chgTask{\egg}) \\
			(\chgAgent{\rightN}, \chgTask{\bowl})
		\end{bmatrix},
        \end{aligned}
\end{equation}
where $\Glact[\sK ]$ denotes the group action on task permutation. Note that since we learn a dedicated policy per subtask, these changes lead to the following group action on the action space of the \gls{pomdp}:
\begin{equation}
	\label{eq:group_action_bimanual}
	\small
	\begin{aligned}
		g_r \!\Glact[\vsA ] \vaa & \!:=\!
		g_r \!\Glact[\vsA ] \!
		\begin{bmatrix}
			\vaa^{\leftN} \sim \pi_{\bowl}(\vo^{\leftN, \bowl}) \\
			\vaa^{\rightN} \sim \pi_{\egg }(\vo^{\rightN, \egg})
		\end{bmatrix}
		\!=\!
            \begin{bmatrix}
			\vaa^{\leftN}
			\sim
			\g_r \Glact[\vsA_{\sN}] (\pi_{\chgTask{\egg }}(
			\vo^{\chgAgent{\rightN},\chgTask{\egg }}))
			\\
			\vaa^{\rightN} \sim \g_r \Glact[\vsA_{\sN}] (\pi_{\chgTask{\bowl}}(
			\vo^{\chgAgent{\leftN},\chgTask{\bowl}})
			)
		\end{bmatrix}
		\!=\!
		\begin{bmatrix}
			\vaa^{\leftN} \sim \pi_{\chgTask{\egg }}(
			\obsFn^{\leftN}(
			\g_r \Glact[\vsS]\vs, \chgTask{\egg })) \\
			\vaa^{\rightN} \sim \pi_{\chgTask{\bowl}}(
			\obsFn^{\rightN}(
			\g_r \Glact[\vsS]\vs, \chgTask{\bowl}))
		\end{bmatrix}
	\end{aligned}
\end{equation}

Essentially, this shows that the action of a robot arm in the reflected environment equals the symmetry-transformed action of the opposite arm in the original environment (see \cref{fig:overview}-left). Here, $(\Glact[\vsA_\sN])$ denotes the group action on an individual arm’s action space. Crucially, the right-hand side of \cref{eq:group_action_bimanual} relies on the $\G$-equivariance of each subtask policy and observation function, ensuring that the global policy is equivariant. Moreover, this analysis directly extends to multi-robot systems with more complex symmetry groups:

\begin{figure}[!t]
	\centering
	\includegraphics[width=\textwidth]{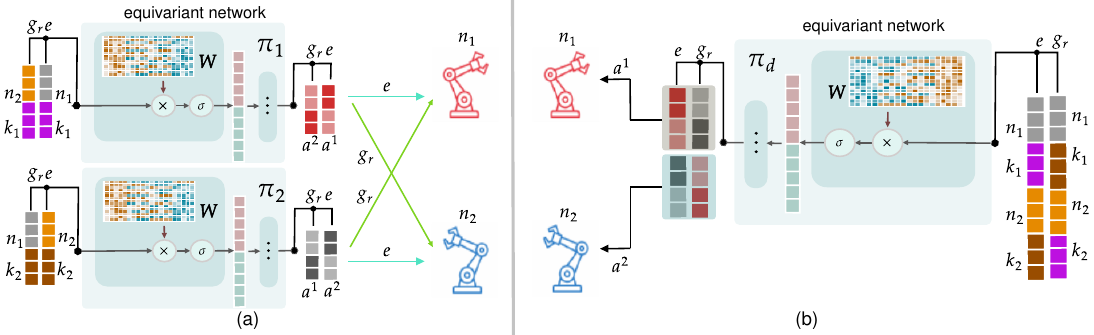}
	\vspace*{-5mm}
	\caption{A comparison of action execution between (a) subtask policies and (b) global policy, where $n_1,n_2$ are agents, $k_1,k_2$ are two subtasks, and $e,g_r$ are transformations in the reflection group.}
        \label{fig:equi_network}
	\vspace*{-5mm}
\end{figure}
\textbf{Morphological symmetries in \acrshort{mtmapomdp}}~~~
Let $( \vsS, \vsA, \vr, \mdpKernel, \initDist, \gamma, \obsSpace, \obsFn, \sK, \sN)$ denote a $N$-robot manipulation \gls{mtmapomdp}, with agents $\sN = \{1,\dots, N\}$, tasks $\sK = \{k_1,\dots, k_N\}$, and agent-task pairs $\sI = \{(1,k_1),\dots\}$ associated with a $\G$-symmetric \gls{pomdp} (as per \cref{def:symm_mdp}). Then, the group action on the \gls{pomdp} action space $\vsA$ is defined via the tensor product (see \cref{note:tensor_product_representation_notation}) of the group actions $(\Glact[\vsA])$ expressed by:
\begin{equation}
	\label{eq:group_action_multi_robot}
	\small
	\begin{aligned}
		g \Glact[\vsA ]
		\vaa
		\!=\!
            g \Glact[\vsA ]
            \begin{bmatrix}
			\vaa^{1} \sim
			\pi_{k_1}(
			\vo^{1, k_1
			})
			\\
			\scalebox{0.5}{$\bm{\vdots}$}
			\\
			\vaa^{N} \sim
			\pi_{k_{N}}(
			\vo^{N,k_{N}
			})
		\end{bmatrix}
		\!=\!
            \begin{bmatrix}
			\vaa^{1} \sim
			\pi_{\chgTask{g \Glact[\sK ] k_1}}(
			\obsFn^{1}(\g \Glact[\vsS]\vs, \chgTask{g \Glact[\sK ] k_1})
			)
			\\
			\scalebox{0.5}{$\bm{\vdots}$}
			\\
			\vaa^{N} \sim
			\pi_{\chgTask{g \Glact[\sK ] k_{N}}}(
			\obsFn^{N}(\g \Glact[\vsS]\vs, \chgTask{g \Glact[\sK ] k_{N}})
			)
		\end{bmatrix}
		\!,
	\end{aligned}
\end{equation}

\Cref{eq:group_action_multi_robot} generalizes the bimanual manipulation example in \cref{eq:group_action_bimanual} to an $N$-robot task with $\G$-equivariant dynamics (see \cref{eq:mdp_dyn_equivariance}). Crucially, this analysis identifies the symmetry constraints for each subtask policy and observation function of the \gls{mtmapomdp} while characterizing the group actions on the global action space of the \gls{pomdp}. This enables us to first learn $\G$-equivariant policies for each \textit{subtask} and then distill them into a global $\G$-equivariant policy for the entire system.

\paragraph{Symmetry-aware learning of subtask policies}
\label{sec:individual}
After decomposing a multi-robot manipulation tasks into subtasks we learn a policy for each. Since each subtask has a unique observation space—comprising the assigned robot arm's state and the task-specific objects state—each subtask policy is parameterized as a $\G$-equivariant \gls{nn} \citep{bronstein2021geometric} (see Fig.~\ref{fig:equi_network}(a)), satisfying:
\begin{equation}
	\label{eq:sub_task_policy_parametrization}
	\small
	g \Glact[\vsA_{\sN}] \pi_{k}^{\nnParams_k}(\vo^{n,k})
	=
	\pi_{k}^{\nnParams_k}(\g \Glact[\vsO_k] \obsFn^{n}(\vs, k))
	=
	\pi_{k}^{\nnParams_k}(\obsFn^{(\g \Glact[\sI ]\sI)[k]}(\g \Glact[\vsS] \vs, k)), \quad \forall\, (n,k) \in \sI, g \in \G.
\end{equation}
Here $\nnParams_k$ are the parameters of the $k$-th subtask network, the group action on the observation space $\Glact[\vsO_k]$ is analogous to $\Glact[\vsA_{\sN}]$ which transforms the robot-arm state and task-specific measurements to its symmetric counterpart, and $\sI[\cdot]$ operator denotes the agent query from the agent-task pair, whose task equals $(\cdot)$. See \citep{ordonez2025morphological} for details on how to construct these actions.

Under the assumption that each subtask reward is $\G$-invariant, i.e.,
$
	\reward_k(\obsFn^{n}(\vs, k))
	=
	\reward_k(\obsFn^{(\g \Glact[\sI ]\sI)[k]}(\g \Glact[\vsS]\vs, k))
$ for all $(n,k)\in\sI,\; g\in\G$
---a premise that holds naturally in dexterous manipulation tasks with morphological symmetries because most reward terms depend on hand–object pose errors--- the corresponding subtask value function can be parameterized by a $\G$-invariant \gls{nn} satisfying:
\begin{equation}
	\label{eq:sub_task_value_parametrization}
	\small
	V_{k}^{\nnParams_k}(\vo^{n,k})
	=
	V_{k}^{\nnParams_k}\!\bigl(\g \Glact[\vsO_k]\obsFn^{n}(\vs, k)\bigr)
	=
	V_{k}^{\nnParams_k}\!\bigl(\obsFn^{(\g \Glact[\sI ]\sI)[k]}(\g \Glact[\vsS]\vs, k)\bigr),
	\quad \forall\, (n,k)\in\sI,\; g\in\G.
\end{equation}
This parameterization allows us to employ the \gls{ppo} algorithm \citep{schulman2017proximal} to learn the $K$ subtask $\G$-equivariant policies and $\G$-invariant value functions in parallel \citep{zinkevich2001symmetry_mdp_implications}.

\textbf{Global $\G$-equivariant policy distillation}~~~
We follow a teacher-student paradigm~\cite{chen2023visual} that distills subtask policies into a global $\G$-equivariant policy—which yields an ambidextrous policy in the case of bimanual manipulation.
Specifically, the learned $N$ subtask policies serve as expert policies to generate a dataset of state–action pairs $\sD = \{(\vs_i, \vaa_i)\}_{i=1}^{M}$ (see \cref{eq:group_action_multi_robot}), which we use to learn a global policy $\pi_d^{\nnParamsDist}$ satisfying:
\begin{equation}
	\label{eq:global_policy}
	\small
	g \Glact[\vsA]\pi_d^{\nnParamsDist}(\obsFn(\vs)) = \pi_d^{\nnParamsDist}(g \Glact[\vsO] \obsFn(\vs)) = \pi_d^{\nnParamsDist}(\obsFn(g \Glact[\vsS] \vs)), \quad \forall\, g \in \G, (\vs,\vaa) \in \vsS \times \vsA.
\end{equation}
Here $\nnParamsDist$ are the network parameters; $\Glact[\vsA]$ and $\Glact[\vsO]$ are the group actions on the global action and observation spaces (see \cref{eq:group_action_multi_robot,app:envs}). Notably, the distilled policy infers task–arm assignments directly from demonstrations (Fig.~\ref{fig:equi_network}(b)), while $\G$-equivariance guarantees identical performance from any symmetric initial state—i.e.\ $\vs_0=\bar{\vs}$ and $\vs_0=g\Glact[\vsS]\bar{\vs}$ yield the same outcome for all $g\!\in\!\G$. This constraint boosts robustness and promotes generalization to unseen configurations \citep{ordonez2025morphological,higgins2022symmetry_deepmind,wang2022robot}. In addition, we ensure that the global policy is trained exclusively on non-privileged observations, enabling zero-shot deployment in the real world (Fig.~\ref{fig:overview}-right). The equivariant/invariant MLPs are implemented using the ESCNN library~\cite{weiler2019general}. 

\textbf{Curriculum Learning for Sim-to-Real Transfer}~~~
To smooth sim-to-real transfer, we employ a simple curriculum that progresses whenever the agent surpasses a success-rate threshold. It comprises two components. \textbf{Randomization}: training begins with scene-level symmetry randomization; once performance stabilizes, environment variations (object poses, physical parameters) are gradually introduced. \textbf{Safety penalty}: collision and energy penalties are phased in later, avoiding early over-constraint that would produce overly conservative, exploration-averse policies.

\begin{figure}[!t]
	\centering
	\includegraphics[width=\textwidth]{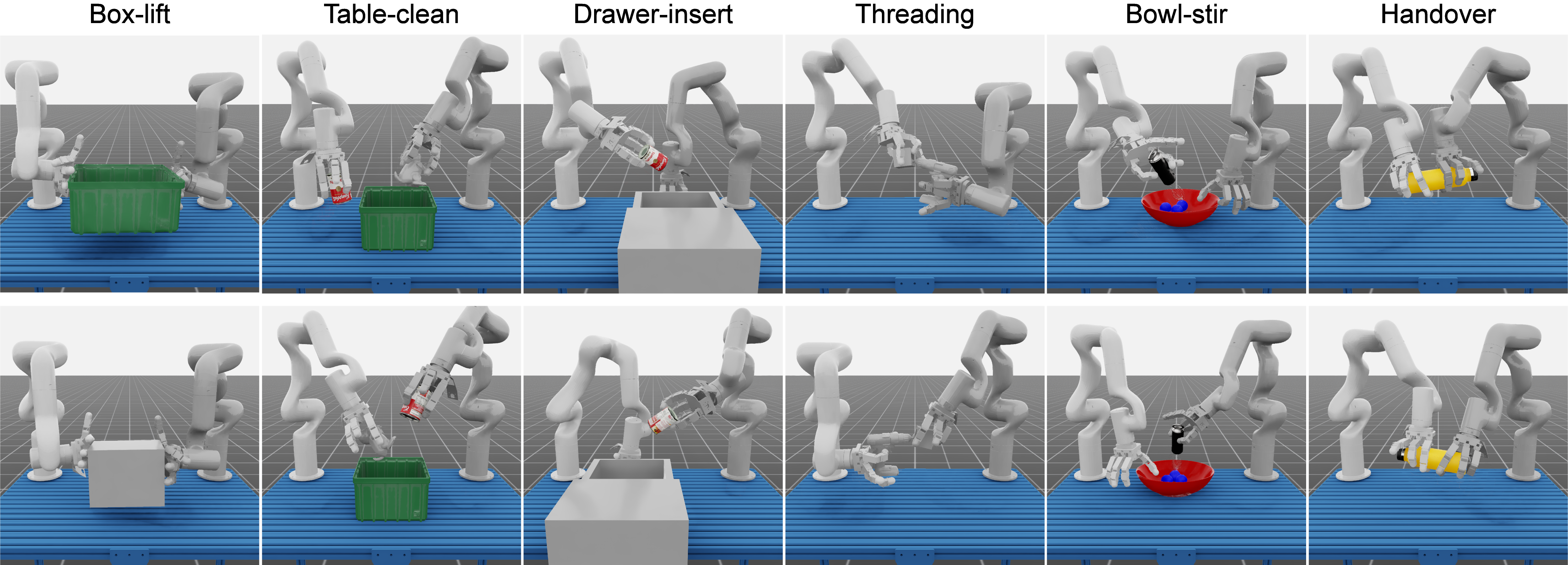}
	\vspace*{-5mm}
	\caption{Our benchmark of six bimanual dexterous manipulation tasks with diverse levels of cooperation and dexterity (TOP); and their symmetric counterparts (Below).}
	\vspace*{-5mm}
        \label{fig:tasks}
\end{figure}

\section{Experiments}
\label{sec:experiments}

We evaluate our method on six simulated bimanual manipulation tasks: \texttt{box-lift}, \texttt{table-clean}, \texttt{drawer-insert}, \texttt{threading}, \texttt{bowl-stir}, and \texttt{handover}. These tasks span a range of coordination and dexterity challenges (see Fig.~\ref{fig:tasks}). We validate the learned policy across all simulated tasks and further deploy it in the real world on two representative tasks: \texttt{box-lift} and \texttt{table-clean}, showcasing effective transfer from simulation to the real world, enabled by curriculum learning. To control the robot, the policy outputs relative joint position actions $a_t$, with joint commands computed as $q^\text{cmd}_{t+1} = q^\text{cmd}_t + \eta \cdot \text{EMA}(a_t)$, where $\eta$ is a scaling factor and EMA denotes exponential moving average smoothing~\citep{chen2024vegetable}. Task setups, success criteria, and reward definitions are detailed in~\Cref{app:envs}, while the full real-world setup is provided in~\Cref{app:real_world}. All simulation experiments are conducted using NVIDIA Isaac Lab~\citep{mittal2023orbit}.

\paragraph{Baselines and Evaluation Metric}
We evaluate five PPO-based baselines, each targeting a specific design aspect: action space dimensionality, task decomposition, value function structure, and symmetry handling via equivariant networks or data augmentation. The baselines include: (1) \gls{eppo}, a single 44-DoF equivariant policy; (2) \gls{ippo}, two independent policies fixed on each arm and trained to handle both subtasks under scene randomization; (3) \gls{eippo}, a 22-DoF single-arm policy shared across arms with task encoding, effectively doubling training data compared to \gls{ippo}; (4) \gls{smc}, our architecture with a centralized value function; and (5) \gls{smaug}, which replaces equivariant networks with on-policy symmetry-based data augmentation~\citep{bao2025toward}. All baselines (and ours) use shared hyperparameters (\Cref{app:hyperparameters}), and a detailed comparison is summarized in \Cref{app:baselines}. Task success rate is the primary metric, averaged over five random seeds with 4096 rollouts each in simulation. Real-world evaluation reports both overall task and per-hand subtask success over 30 independent trials.

\begin{figure}[t!]
	\centering
	\includegraphics[width=0.92\textwidth]{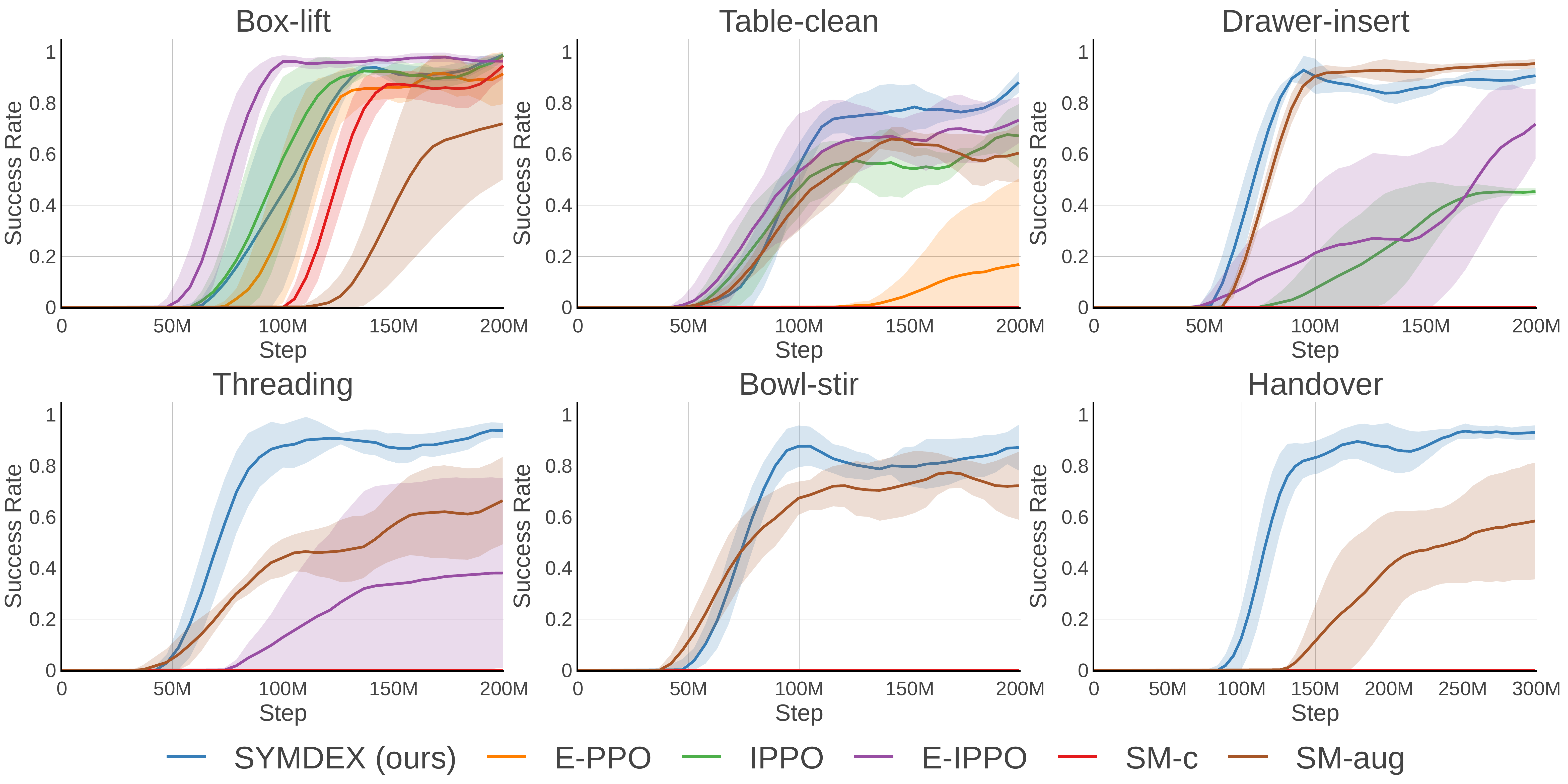}
	\vspace*{-1.5mm}
	\caption{
		Performance of \textsc{SYMDEX} and baseline methods on six benchmark tasks. \textsc{SYMDEX} consistently learns all six tasks and achieves success rates exceeding 80\%, outperforming all baselines.}
	\vspace*{-5mm}
	\label{fig:performance}
\end{figure}



\subsection{Simulation Results}
We evaluate \textsc{SYMDEX} on our simulation benchmark against all baselines, where symmetry transformations are randomly applied to the initial state. As shown in Fig.~\ref{fig:performance}, \textsc{SYMDEX} consistently learns all six tasks with success rates exceeding 80\%, significantly outperforming the baselines.

\textbf{Advantage of Task Decomposition}~~~Task decomposition is highly beneficial when the subtasks assigned to each arm differ significantly. For example, the baseline \acrshort{eppo}, which jointly controls the entire system (44 DoF, cf. Tab.~\ref{tab:baseline}), succeeds only on \texttt{box-lift}, partially on \texttt{table-clean}, and fails on the rest. This occurs because in \texttt{box-lift} and \texttt{table-clean}, both arms perform similar actions, making joint learning tractable, whereas when arm subtasks diverge, \acrshort{eppo}'s monolithic policy struggles to specialize appropriately.

Moreover, decomposing the task at the subtask level—in addition to at the robot arm level—is critical. Baselines like \acrshort{ippo} and \acrshort{eippo} use a decomposed 22 DoF action space, yet each policy remains need to select and perform both subtasks, i.e., a multi-task policy. While \acrshort{ippo} and \acrshort{eippo} perform comparably to \textsc{SYMDEX} on \texttt{box-lift} and \texttt{table-clean}, they fail to generalize to other tasks. In contrast, \textsc{SYMDEX} assigns a policy per subtask, avoiding the issues of multi-task learning.


\textbf{Impact of $\G$-equivariance/invariance Constraints}~~~ When comparing \textsc{SYMDEX} to \acrshort{smaug}---which employs on-policy data augmentation \citep{bao2025toward}---, our proposed method consistently outperforms across all tasks (see \cref{fig:performance}). A similar trend is observed when comparing \acrshort{ippo} and \acrshort{eippo}. This performance boost across all tasks highlights the advantage of embedding symmetry priors directly into the network architecture rather than relying solely on data augmentation.

\textbf{Impact of Centralized Learning}~~~We examine the effect of global value functions on multi-arm cooperation. Both \acrshort{eppo} and \acrshort{smc} use global critics to estimate total rewards across subtasks; however, our results show that such designs suffer from poor credit assignment in complex, contact-rich settings. Notably, \acrshort{eppo} outperforms \acrshort{smc} on \texttt{box-lift} and \texttt{table-clean}, despite \acrshort{smc}'s reduced action space via task decomposition. This suggests that while decomposition lowers dimensionality, it can introduce uncertainty in joint optimization, hindering accurate reward assignment. Our findings indicate that deglobal value functions, as used in SYMDEX, are more effective for high-dimensional, coordinated manipulation tasks.

\textbf{Distillation}~~~We use a teacher-student distillation approach to train a unified global policy (\cref{sec:method_intro}). We compare three student variants: a vanilla Gaussian policy, an equivariant Gaussian policy, and an equivariant diffusion policy~\cite{chi2023diffusion}.
As shown in \cref{tab:merged}, both $\G$-equivariant Gaussian and diffusion policies outperform the vanilla Gaussian policy across all six tasks. This suggest that incorporating equivariant constraints facilitates robust policy distillation. Interestingly, the equivariant Gaussian policy performs comparably to the diffusion variant—likely because the teacher policies used for data collection are Gaussian, allowing the Gaussian policy to fit the dataset effectively.

\begin{table}[t!]
	\centering
	\small
	\begin{adjustbox}{max width=\textwidth}
		\begin{tabular}{lllllll}
			\toprule
			Method                 & Box                                & Table                              & Drawer                             & Threading                          & Bowl                               & Handover                           \\
			\midrule
			Gaussian policy (GP)   & $0.83 \pm{0.03}$                   & $0.74 \pm{0.05}$                   & $0.69 \pm{0.09}$                   & $0.62 \pm{0.13}$                   & $0.75 \pm{0.12}$                   & $0.54 \pm{0.23}$                   \\
			Equi. GP               & $0.89 \pm{0.01}$                   & $0.83 \pm{0.01}$                   & $\textbf{0.87} \pm{\textbf{0.07}}$ & $\textbf{0.63} \pm{\textbf{0.17}}$ & $0.87 \pm{0.08}$                   & $\textbf{0.86} \pm{\textbf{0.12}}$ \\
			Equi. Diffusion policy & $\textbf{0.91} \pm{\textbf{0.04}}$ & $\textbf{0.84} \pm{\textbf{0.02}}$ & $\textbf{0.87} \pm{\textbf{0.13}}$ & $0.60 \pm{0.1}$                    & $\textbf{0.88} \pm{\textbf{0.15}}$ & $0.68 \pm{0.18}$                   \\
			\bottomrule
		\end{tabular}
	\end{adjustbox}
	\vspace{0.4cm}
	\begin{adjustbox}{max width=\textwidth}
		\begin{tabular}{lcccccc}
			\toprule
			                        & \multicolumn{3}{c}{Box}            & \multicolumn{3}{c}{Table}                                                                                                                                                              \\
			\cmidrule(lr){2-4} \cmidrule(lr){5-7}
			                        & Subtask 1                          & Subtask 2                          & Overall                            & Subtask 1                          & Subtask 2                          & Overall                            \\
			\midrule
			Equi. GP w/o Curriculum & $0.2 \pm{0.12}$                    & $0.17 \pm{0.23}$                   & $0.13 \pm{0.08}$                   & $0.13 \pm{0.05}$                   & $0.1 \pm{0.08}$                    & $0.07 \pm{0.12}$                   \\
			Equi. GP                & $\textbf{0.87} \pm{\textbf{0.08}}$ & $\textbf{0.83} \pm{\textbf{0.11}}$ & $\textbf{0.77} \pm{\textbf{0.09}}$ & $\textbf{0.83} \pm{\textbf{0.13}}$ & $\textbf{0.67} \pm{\textbf{0.32}}$ & $\textbf{0.63} \pm{\textbf{0.25}}$ \\
			Equi. Diffusion Policy  & $0.7 \pm{0.20}$                    & $0.73 \pm{0.13}$                   & $0.6 \pm{0.23}$                    & $0.73 \pm{0.15}$                   & $0.47 \pm{0.34}$                   & $0.4 \pm{0.21}$                    \\
			\bottomrule
		\end{tabular}
	\end{adjustbox}
	\caption{
		(Top) Simulation distillation results for six different tasks using three architectural choices. (Below) Real-world performance comparison on \texttt{box-lift} and \texttt{table-clean}, assessed through qualitative evaluations by human operators.
	}
	\label{tab:merged}
	\vspace*{-0.2cm}
\end{table}

\subsection{Real-World Results}
We conduct sim-to-real experiments to evaluate the performance of our distilled policy and its two variants on two real-world tasks: \texttt{box-lift} and \texttt{table-clean}. The real-world setup is in Fig.~\ref{fig:real-world-figure} and videos are in supplementary material. As shown in~\Cref{tab:merged}-Bottom, the equivariant Gaussian policy consistently outperforms both its counterpart trained without curriculum and the variant that replaces the Gaussian model with a diffusion model.

First, we observe that removing the curriculum leads to a significant performance drop, highlighting the importance of domain randomization and safety constraints for successful sim-to-real transfer. Second, although the equivariant diffusion policy achieves better distillation results than the Gaussian policy in simulation, the Gaussian policy proves to be more robust in the real world. We attribute this to the homogeneous dataset collected from the teacher policies: the diffusion model struggles to generalize to out-of-distribution observations, particularly under imperfect state estimation from the perception system. In contrast, the Gaussian policy directly fits the teacher policy’s distribution, making it more robust to the sim-to-real gap.
\subsection{Scaling to Multi-Robot Symmetry}
\label{sec:scalability_multi-robot}
\begin{wrapfigure}{r}{0.25\textwidth}
	\vspace*{-0.2cm}
	\centering
	\includegraphics[width=0.95\linewidth]{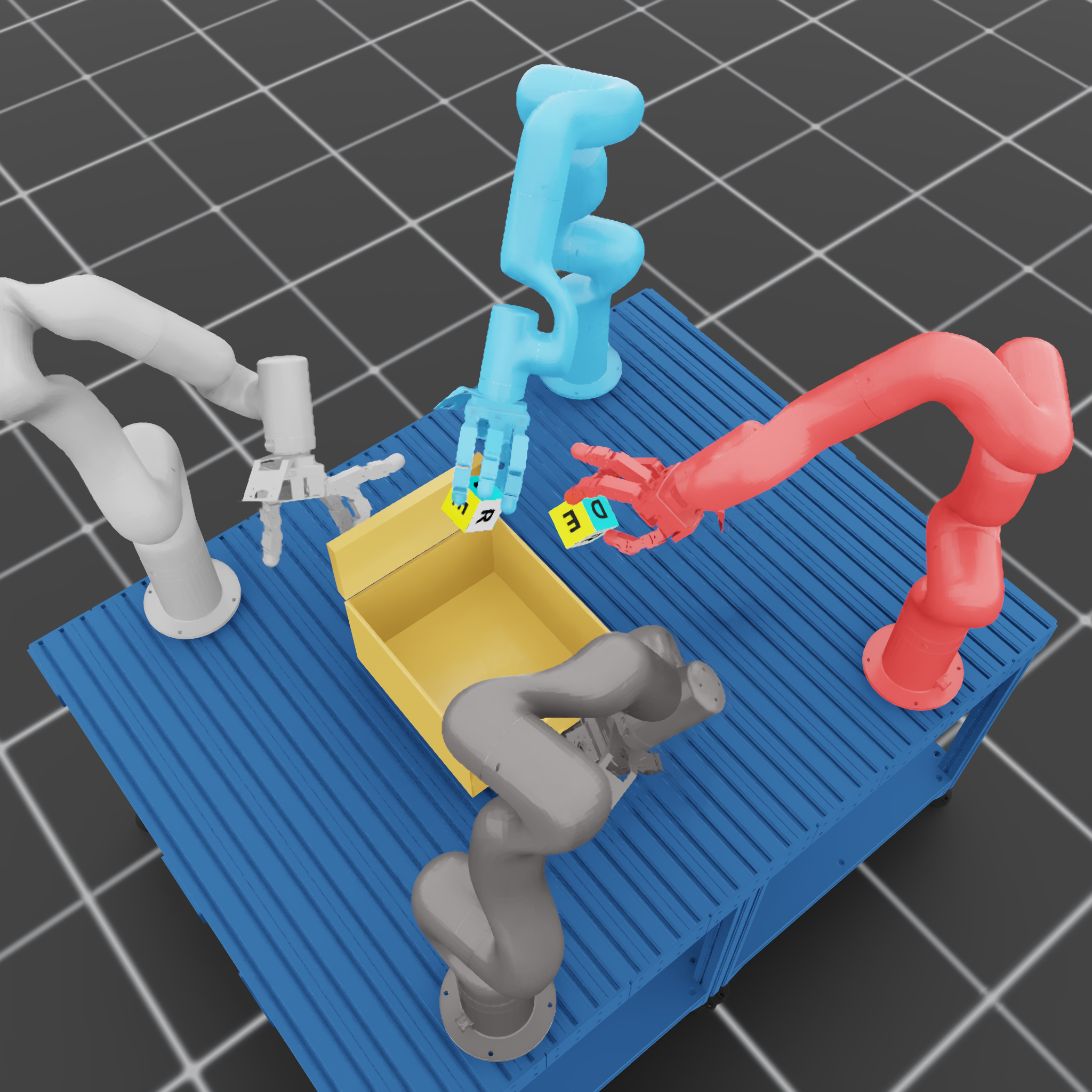}
	\caption{The four-arm system setup.}
	\label{fig:4_arm_single}
	\vspace*{-0.4cm}
\end{wrapfigure}
We demonstrate the scalability of \textsc{SYMDEX} on a multi-robot task involving a system of four arms,
each equipped with a right dexterous hand. The objective is for two arms to hold the flaps of a cardboard box while the other two arms pick up objects from the table and place them into the box (Fig.~\ref{fig:4_arm_single}). Once the objects are inside, the two arms holding the flaps then close the box. Since a constant force is applied to the box flaps to keep them open, the task requires coordinated collaboration among all arms to succeed.

The four-arm system exhibits symmetry under the group $\G=\CyclicGroup[4]=\{e,g_r,g_r^2,g_r^3\mid g_r^4=e\}$, where $g_r$ is a $90^\circ$ rotation about the vertical axis. Following the method described in Sec.~\ref{sec:method_intro}, we treat each arm as an agent and assign a specific subtask to each. After training, the system successfully completes the task from different orientations, with Fig.~\ref{fig:4-arm} visualizing all symmetric scenarios from a fixed camera viewpoint. Additional experiment results are provided in~\Cref{app:additional_exp}.

\section{Conclusion}
In this work, we presented SYMDEX, a novel \gls{rl} framework for learning morphological symmetry-aware policies that achieve ambidextrous bimanual manipulation. \textsc{SYMDEX} enables efficient policy learning across six complex dexterous manipulation tasks, enhances policy robustness through symmetry exploitation, and achieves zero-shot sim-to-real transfer on two real-world tasks. Furthermore, we demonstrated the scalability of \textsc{SYMDEX} on a four-arm setup, successfully handling more intricate symmetry groups and multi-agent coordination. We believe that incorporating symmetry as an inductive bias offers a powerful tool for advancing robotic learning, particularly as morphologically inspired humanoid and multi-armed robots become increasingly prominent.

\clearpage
\section{Limitations}
We acknowledge that the primary limitation of \textsc{SYMDEX} is its reliance on the presence of morphological symmetry within the robotic system. However, we emphasize that such symmetry is common in many modern robotic platforms, including bimanual systems~\citep{huang2023dynamic, lin2024twisting}, tri-arm robots like Trifinger~\citep{wuthrich2020trifinger}, and humanoid robots~\citep{lin2025sim,ze2024generalizable}.

We note that \textsc{SYMDEX} primarily leverages kinematic-level symmetry, where joint positions and end-effector poses are symmetric under group transformation. This design choice allows us to use joint position control, which is sufficient for many manipulation tasks and avoids the need for full dynamic symmetry, as required by torque control. Achieving symmetry at the dynamics level—particularly under reflection—would require the robot components to be true mirror models. While this condition holds for the left and right hands, it does not strictly apply to the arms, which are typically identical in construction rather than mirrored. As a result, their dynamic properties, such as mass distribution and collision avoidance, may not fully follow reflectional symmetry. However, since \textsc{SYMDEX} operates at the kinematic level, this does not significantly impact its effectiveness in practice.

Regarding the failure cases in the real world experiments, we observe that the major issue comes from the perception part. Since our policy is state-based, it depends on accurate multi-object pose tracking, which is difficult in practice. However, our equivariant architecture can also be applied to vision-based inputs, such as RGB-D images and point clouds \cite{deng2021vector}, to improve robustness, which we will leave as future work.

\acknowledgments{This research work has received funding from the Emmy Noether Programme DFG Project Nr. CH 2676/1-1 and the Project Pose Confidences for Telerobotics from Honda Research Institute, Europe.}


\bibliography{references}  

\newpage
\appendix

\part*{Appendix}
\section{Background on group and representation theory}
\label{sec:group_theory}

\subsection*{Group actions and representations}

This section provides a brief overview of the fundamental concepts in group and representation theory, which are used to define symmetry groups of robotic systems and \glspl{mdp}. For a comprehensive and intuitive background on group and representation theory in machine learning, we refer the reader to \citet{weiler2023EquivariantAndCoordinateIndependentCNNs}.

To begin, we define a group as an abstract mathematical object.
\begin{definition}[Group] \label{def:group}
	A group is a set $\G$, endowed with a binary composition operator defined as:
	\begin{subequations}
		\begin{equation}
			\mapping
			{(\Gcomp)}
			{\G \times \G}{\G}
			{(\g_1, \g_2)}
			{\g_1 \Gcomp \g_2,}
			\label{eq:group_operation}
		\end{equation}
		such that the following axioms hold:
		\begin{align}
			\text{Associativity:} & \quad (\g_1 \Gcomp \g_2) \Gcomp \g_3 = \g_1 \Gcomp (\g_2 \Gcomp \g_3), \quad \forall\; \g_1,\g_2,\g_3 \in \G, \label{eq:group_associativity}      \\
			\text{Identity:}      & \quad \exists\; e \in \G \; \text{such that} \; e \Gcomp \g = \g = \g \Gcomp e, \quad \forall\; \g \in \G, \label{eq:group_identity}              \\
			\text{Inverses:}      & \quad \forall\; \g \in \G, \; \exists\; \g^{-1} \in \G \; \text{such that} \; \g \Gcomp \g^{-1} = e = \g^{-1} \Gcomp \g. \label{eq:group_inverse}
		\end{align}
	\end{subequations}
\end{definition}
We focus on symmetry groups—that is, groups of transformations acting on a set $\vsX$ where each transformation is a bijection preserving an intrinsic property. For example, if $\vsX$ represents the states of a dynamical system, the invariant property might be the state's energy (see \cref{fig:overview}).
\begin{definition}[Group action on a set \citep{weiler2023EquivariantAndCoordinateIndependentCNNs}]
	\label[definition]{def:left_group_action}
	Let $\vsX$ be a set endowed with the symmetry group $\G$. The (left) group action of the group $\G$ on the set $\vsX$ is a map:
	\begin{subequations}
		\begin{align}
			\mapping
			{(\Glact)}
			{ \G \times \vsX}{\vsX}
			{(\g,\vx)}
			{\g \Glact \vx}
		\end{align}
		that is compatible with the group composition and identity element $e \in \G$, such that the following properties hold:
		\begin{align}
			\text{Identity:}      & \quad e \Glact \vx = \vx,                                            & \quad \forall\; \vx \in \vsX \label{eq:symmetry_action_on_set_identity}                                   \\
			\text{Associativity:} & \quad (\g_1 \Gcomp \g_2) \Glact \vx = \g_1 \Glact (\g_2 \Glact \vx), & \quad \forall\; \g_1, \g_2 \in \G, \forall\; \vx \in \vsX \label{eq:symmetry_action_on_set_associativity}
		\end{align}
	\end{subequations}
\end{definition}

We are primarily interested in studying symmetry transformations on sets with a vector space structure. In most practical cases, the group action on a vector space is linear, allowing symmetry transformations to be represented as linear invertible maps. These maps can be expressed in matrix form once a basis for the space is chosen.

\begin{definition}[Linear group representation]
	\label{def:group_representation}
	Let $\vsX$ be a vector space endowed with the symmetry group $\G$. A \emph{linear representation} of  $\G$ on $\vsX$ is a map, denoted by $\rep[\vsX]$, between symmetry transformation and invertible linear maps on $\vsX$ (i.e., elements of the general linear group $\GLGroup(\vsX)$):
	\begin{subequations}
		\begin{equation}
			\mapping{\rep[\vsX]}{\G}{\GLGroup(\vsX)}{\g}{\rep[\vsX](\g),}
		\end{equation}
		such that the following properties hold:
		\begin{align}
			\text{composition} & : \rep[\vsX](\g_1 \Gcomp \g_2) = \rep[\vsX](\g_1) \rep[\vsX](\g_2),
			                   & \quad \forall\; \g_1, \g_2 \in \G,
			\label{eq:group_representation_composition}                                                                                                                   \\
			\text{inversion}   & : \rep[\vsX](\g^{-1}) = \rep[\vsX](\g)^{-1}, \quad                  & \forall\; \g \in \G. \label{eq:group_representation_invertibility} \\
			\text{identity}    & :  \rep[\vsX](g \Gcomp \g^{-1}) = \rep[\vsX](e) = \mI,              &
			\label{eq:group_representation_identity}
		\end{align}
		\label{eq:group_representation_properties}
		\noindent
		Whenever the vector space is of finite dimension $|\vsX| = n < \infty$, linear maps admit a matrix form  $\rep[\vsX](\g) \in \R^{n \times n}$, once a basis set $\bSet{\vsX}$ for the vector space $\vsX$ is chosen. In this case, \cref{eq:group_representation_composition,eq:group_representation_invertibility,eq:group_representation_identity} show how the composition and inversion of symmetry transformations translate to matrix multiplication and inversion, respectively. Moreover, $\rep[\vsX]$ allows to express a (linear) group action (\cref{def:left_group_action}) as a matrix-vector multiplication:
		\begin{align}
			\mapping
			{(\Glact)}
			{ \G \times \vsX}{\vsX}
			{(\g,\vx)}
			{\g \Glact \vx := \rep[\vsX](\g) \vx}
		\end{align}
	\end{subequations}
\end{definition}

\begin{definition}[Tensor product representation]
	\label{def:tensor_product_representation}
	Let $\vsX$ and $\vsY$ be (finite-dimensional) vector spaces endowed with a common symmetry group $\G$.  
	Denote by
	$
		\rep[\vsX] : \G \to \GLGroup(\vsX),
	$ and $
		\rep[\vsY] : \G \to \GLGroup(\vsY)
	$
	the corresponding linear representations.  
	The tensor product representation is defined through the Kronecker product of the representations of group actions on the vector spaces:
	\begin{equation}
		\mapping
		{(\rep[\vsX] \otimes \rep[\vsY])}
		{\G}{\GLGroup(\vsX \otimes \vsY)}
		{\g}
		{\rep[\vsX](\g)\, \otimes\, \rep[\vsY](\g),}
		\label{eq:tensor_rep_mapping}
	\end{equation}
	\begin{note}
		\label{note:tensor_product_representation_notation}
		Whenever denoting group actions by $(\Glact[\vsX])$ and $(\Glact[\vsY])$, we will use the notation $\Glact[\vsX \otimes \vsY]$ to denote the group action on the tensor product space $\vsX \otimes \vsY$. Such that:
	\begin{equation}
		\mapping{\Glact[\vsX \otimes \vsY]}
		{\G \times (\vsX \otimes \vsY)}
		{(\vsX \otimes \vsY)}
		{(\g, \vx \otimes \vy)}
		{[\rep[\vsX](\g) \otimes \rep[\vsY](\g)] (\vx \otimes \vy)}
	\end{equation}

	\end{note}
\end{definition}

\subsection*{Maps between symmetric vector spaces} \label{sec:group_equivariant_maps}

We will frequently study and use linear and non-linear maps between symmetric vector spaces. Our focus is on maps that preserve entirely or partially the group structure of the vector spaces. These types of maps can be classified as $\G$-equivariant, $\G$-invariant maps:

\begin{definition}[$\G$-equivariant and $\G$-invariant maps]
	\label{def:equivariantMaps}
	Let $\vsX$ and $\vsY$ be two vector spaces endowed with the same symmetry group $\G$, with the respective group actions $\Glact_\vsX$ and $\Glact[\vsY]$. A map $f: \vsX \mapsto \vsY$ is said to be $\G$-equivariant if it commutes with the group action, such that:
	\begin{subequations}
		\begin{equation}
			\begin{split}
				\g \Glact[\vsY] \vy = \g \Glact[\vsY] f(\vx) &= f (\g \Glact[\vsX] \vx), \quad \forall \vx \in \vsX, \g \in \G. \\
				\rep[\vsY](\g) f(\vx) &= f(\rep[\vsX](\g) \vx)
			\end{split}
			\qquad \vcenter{\hbox{$\iff$}}
			\qquad
			\vcenter{\hbox{
					\homomorphismDiag
					{\vsX}
					{\vsY}
					{\Glact[\vsX]}
					{\Glact[\vsY]}
					{f}
				}}
		\end{equation}

		A specific case of $\G$-equivariant maps are the $\G$-invariant ones, which are maps that commute with the group action and have trivial output group actions $\Glact[\vsY]$ such that $\rep[\vsY](\g) = \mI$ for all $\g \in \G$. That is:
		\begin{equation}
			\label{eq:invariant_maps}
			\begin{split}
				\vy = \g \Glact[\vsY] f(\vx) &= f(\g \Glact[\vsX] \vx), \quad \forall \vx \in \vsX, \g \in \G. \\
				\vy = \rep[\vsY](\g) f(\vx) & = f(\rep[\vsX](\g) \vx)
			\end{split}
			\qquad \vcenter{\hbox{$\iff$}}
			\qquad
			\vcenter{\hbox{
					\invariantDiag
					{\vsX}
					{\vsY}
					{\Glact[\vsX]}
					{f}
					{\Glact[\vsY]}
				}}
		\end{equation}
	\end{subequations}
\end{definition}

\section{Symmetries in \glspl{mdp}}

This section introduces a formal definition and notation of symmetries in \glspl{pomdp}, based on the previous works of \citep{ordonez2025morphological,zinkevich2001symmetry_mdp_implications,van2020mdp}.

\begin{definition}[Symmetric \gls{pomdp}]
	\label{def:symm_mdp_appx}
	A \gls{pomdp} $(\vsS, \vsA, r, \mdpKernel, \initDist, \gamma, \obsSpace, \obsFn)$ possess the symmetry group $\G$ when the state and action spaces $\vsS$ and $\vsA$ admit group actions $(\Glact[\vsS])$ and $(\Glact[\vsA])$, and $(r, \mdpKernel, \initDist)$ are all $\G$-invariant. That is, if for every $\g\in\G$, $s,s'\in\vsS$, and $a\in\vsA$, we have:
	\begin{equation}
		\label{eq:symm_mdp_appx}
		\small
		\mdpKernel(\g \Glact[\vsS] s' \mid \g \Glact[\vsS] s, \g \Glact[\vsA] a) = \mdpKernel(s' \mid s, a),
		\quad\;\;\;
		\initDist(\g \Glact[\vsS] s) = \initDist(s),
		\quad\;\;
		r(\g \Glact[\vsS] s, \g \Glact[\vsA] a) = r(s, a).
	\end{equation}
	\gls{pomdp}'s satisfying \cref{eq:symm_mdp} are constrained to have \textbf{optimal} policy and value functions satisfying:
	\begin{equation}
		\label{eq:symm_mdp_sol_appx}
		\small
		\ubcolor{
			\g \Glact[\vsA] \pi^*(\obsFn(s)) = \pi^*(\obsFn(\g \Glact[\vsS] s))}
		{\text{\tiny Policy $\G$-equivariance}},
		\qquad
		\ubcolor[awesomeorange]{
			V^{*}(\obsFn(s)) = V^{*}(\obsFn(\g \Glact[\vsS] s))
		}{\text{\tiny Value function $\G$-invariance}}
		,
		\quad
		\forall\; s \in \vsS,\;\g \in \G. \;
		\text{(refer to \citep{zinkevich2001symmetry_mdp_implications})}
	\end{equation}
\end{definition}

\begin{proposition}[Conditions for optimality \citep{ordonez2023dynamics}]
    \label{prop:conditions_for_optimality_appx}
    Given the $\G$-equivariance constraint on the \textbf{optimal} policy $\pi^*$ and the $\G$-invariance of the optimal value function $V^*$ in \cref{eq:symm_mdp_sol_appx} of a symmetric \gls{pomdp}, any parametric policy $\pi_{\nnParams}\colon \vsO \to \vsA$ and value function $V_{\nnParams}\colon \vsO \to \mathbb{R}$ can be made $\G$-equivariant and $\G$-invariant, respectively, if the observation function $\obsFn$ is $\G$-equivariant, thus endowing the observation space with the same symmetry group $\G$ and group action $\Glact[\obsSpace]$.

    This holds because for the composition of two functions to be $\G$-equivariant $
	(
		\pi_{\nnParams} \circ \obsFn\colon \vsS \to \vsA
	)$ or $\G$-invariant $ 
    (
    V_{\nnParams} \circ \obsFn\colon \vsS \to \mathbb{R}
    )$, both functions must be $\G$-equivariant, such that:
    \begin{align}
        \label{eq:equivariance_condition_appx}
        \small
        \g \Glact[\vsA] \pi_{\nnParams}(\obsFn(\vs)) 
        &= 
        \pi_{\nnParams}(\g \Glact[\vsO] \obsFn( \vs))
        &&=
        \pi_{\nnParams}(\obsFn(\g \Glact[\vsS] \vs)),
        &&&
        \quad
        \forall\; \vs \in \vsS,\;\g \in \G.
        \\
        V_{\nnParams}(\obsFn(\vs)) 
        &=
        V_{\nnParams}(\g \Glact[\vsO] \obsFn(\vs)) 
        &&=
        V_{\nnParams}(\obsFn(\g \Glact[\vsS] \vs)),
        &&&
    \end{align}
\end{proposition}

\section{Related Work} \label{sec:related_work}

\paragraph{Symmetry in robotic manipulation} 
Recent works leverage inherent rotational symmetries in 3D environments to design $\SE$-, $\SO$-, or $\SO[2]$-equivariant grasping and pose estimation pipelines \citep{ryu2024diffusion,brehmer2023edgi,huang2022equivariant,zhu2022sample,wang2022mathrm,wang2020incorporating,wang2024equivariant,hu2025push}. These approaches typically define the \gls{mdp}'s action as the target task-space configuration and use off-the-shelf \gls{ik} solvers with built-in collision avoidance. In contrast, our method focuses on multi-robot manipulation environments with the action space defined in generalized coordinates, forcing the policy to implicitly learn collision avoidance, in-hand manipulation, and \gls{ik}. Furthermore, our work focuses on leveraging the morphological symmetries \citep{ordonez2025morphological} of the manipulation \gls{mdp}, rather than the environmental symmetries of Euclidean space. Consequently, learned policies are equivariant only to \textit{finite} subgroups of $\EG$, because practical manipulation environments rarely exhibit full $\EG$-symmetry—joint limits and workspace obstacles break the symmetries of the continuous group (see \Cref{def:symm_mdp}).

\paragraph{Morphological symmetry in reinforcement learning}
Considering morphological symmetry priors as an inductive bias is a trend in state-of-the-art robotics research to enhance sample efficiency and policy generalization. There are two main approaches to leverage the symmetry priors of \Cref{eq:symm_mdp_sol}, namely employing equivariant network and data augmentation~\citep{ordonez2025morphological,zinkevich2001symmetry_mdp_implications,van2020mdp}. We studied both methods in our experiments and demonstrated the superior performance of equivariant network when the symmetry is properly defined. However, existing works focus on quadrupedal locomotion~\citep{su2024leveraging,mittal2024symmetry,ordonez2022adaptable,abdolhosseini2019learning}, and in our work we investigate bimanual (and multi-arm) dexterous manipulation. 

\paragraph{Reinforcement learning for (bimanual) dexterous manipulation}
Bimanual dexterous manipulation is a well-known challenging problem in robotics. Recent works focus on specific tasks, underscoring the problem’s complexity. For example, \cite{lin2024twisting} addresses unscrewing a lid, while \cite{huang2023dynamic} and \cite{lan2023dexcatch} focus on handover/catch scenarios between arms. Notably, \cite{lan2023dexcatch} presents simulation-only results, and both \cite{lin2024twisting} and \cite{huang2023dynamic} simplify the system by reducing \gls{dof}—\cite{lin2024twisting} fixes the dual arms and controls only the hands, and \cite{huang2023dynamic} locks several arm joints—thus shrinking the exploration space and avoiding the task’s complexity. In contrast, our work maintains full control over all \glspl{dof} in both arms and hands, preserving the inherent richness—and challenge—of the original problem.

\paragraph{Sim-to-real Transfer} A key challenge is transferring trained policies to the real world. Two primary strategies have emerged for sim-to-real transfer. Teacher-student distillation has been successfully applied in dexterous manipulation \citep{chen2023visual,chen2024vegetable,qi2023hand}. This approach leverages privileged simulation information to teach a student policy that operates under realistic sensory constraints; our method builds on this by incorporating permutation invariance during distillation. The second strategy, curriculum learning, automatically increases task difficulty to improve both generalization and policy robustness~\citep{tiboni2023domain,akkaya2019solving}. For example, in \citep{tiboni2023domain}, it directly maximizes the entropy of the environment distribution as long as the the success rate is sufficiently high. We use a similar idea, but simplifiy the maximum entropy objective to a fixed step curriculum.
\newpage

\section{Pseudoalgorithm}
\label{app:pseudocode}
\begin{algorithm}[H]
\caption{Symmetric Dexterity (SYMDEX)}
\label{algo:symdex}
\begin{algorithmic}[1]
    \STATE \textbf{Input}: number of agents and tasks $N$, initial policies $\{\pi_k\}_{k=1}^N$, initial value functions $\{V_k\}_{k=1}^N$, horizon length $T$, update-to-data (UTD) ratio $G$.
    \FOR{each iteration}
        \FOR{$t = 1, \cdots, T$}
            \STATE Observe state $\vs_t$ and construct observation $\vo_t = \obsFn(\vs_t, \sI_k)$.
            \STATE Sample action $\{\vaa_t^n \sim \pi_{\sI_{k_n}}(\vo^{n_t,\sI_{k_n}})\}_{n=1}^N$ for each agent-task pair.
            \STATE Concatenate for global action $\vaa_t = \oplus_{n \in \sN} \vaa_t^{n}$.
            \STATE Execute action $\vaa_t$ in the environment and collect data $\{(\vo_t^{n,\sI_{k_n}}, \vaa_t^n, \vr_t^{n, \sI_{k_n}}, \vo_{t+1}^{n,\sI_{k_n}})\}_{n=1}^N$.
        \ENDFOR{}
        \STATE Compute advantage estimates $\{\bm{\Lambda}^n\}_{n=1}^N$ using $V_{\sI_{k_n}}$.
        \FOR{$g = 1, \cdots, G$}
            \FOR{$n = 1, \cdots, N$}
                \STATE Sample a batch $B_g$ from $\{(\vo_t^{n,\sI_{k_n}}, \vaa_t^n, \vr_t^{n, \sI_{k_n}}, \vo_{t+1}^{n,\sI_{k_n}})\}$.
                \STATE Update policy $\pi_{\sI_{k_n}}$ on \gls{ppo} loss.
                \STATE Update value function $V_{\sI_{k_n}}$ on MSE loss. 
            \ENDFOR
        \ENDFOR{}
    \ENDFOR{}
    \end{algorithmic}
\end{algorithm}

\section{Environment Details}
\label{app:envs}
In this section, we provide a detailed description for all six tasks, including task descriptions, success criteria, and reward functions. For all tasks, the episode length is $100$.

\begin{minipage}{0.3\textwidth}
\centering
\includegraphics[width=0.65\linewidth]{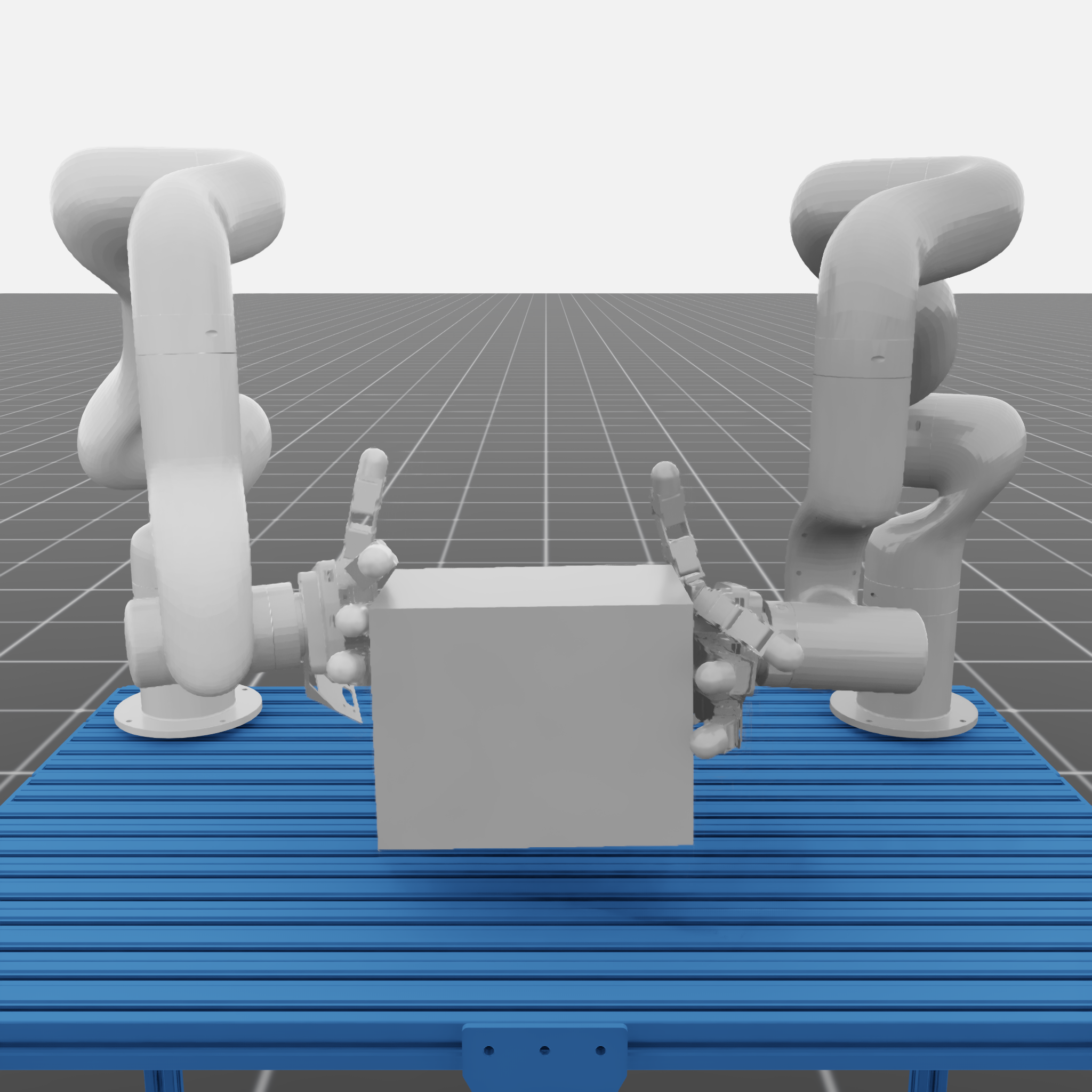}
\vspace{0.8em}
\end{minipage}
\begin{minipage}{0.7\textwidth}
\texttt{Box-lift}: The goal is to use both hands to lift a box and hold it at a target pose. Each subtask involves one hand approaching the box from one side and lifting it in coordination with the other hand. The two subtasks are identical but mirrored, requiring tight cooperation between both agents.
\end{minipage}
\textbf{Success criteria}: The box must be held at the target pose for 20 consecutive steps.\\
\textbf{Reward functions for both subtasks}: (1) A hand alignment reward that encourages the palm to align with the side of the box; (2) A box pose reward that encourages the box’s position and orientation to match the target.
\\

\begin{minipage}{0.3\textwidth}
\centering
\includegraphics[width=0.65\linewidth]{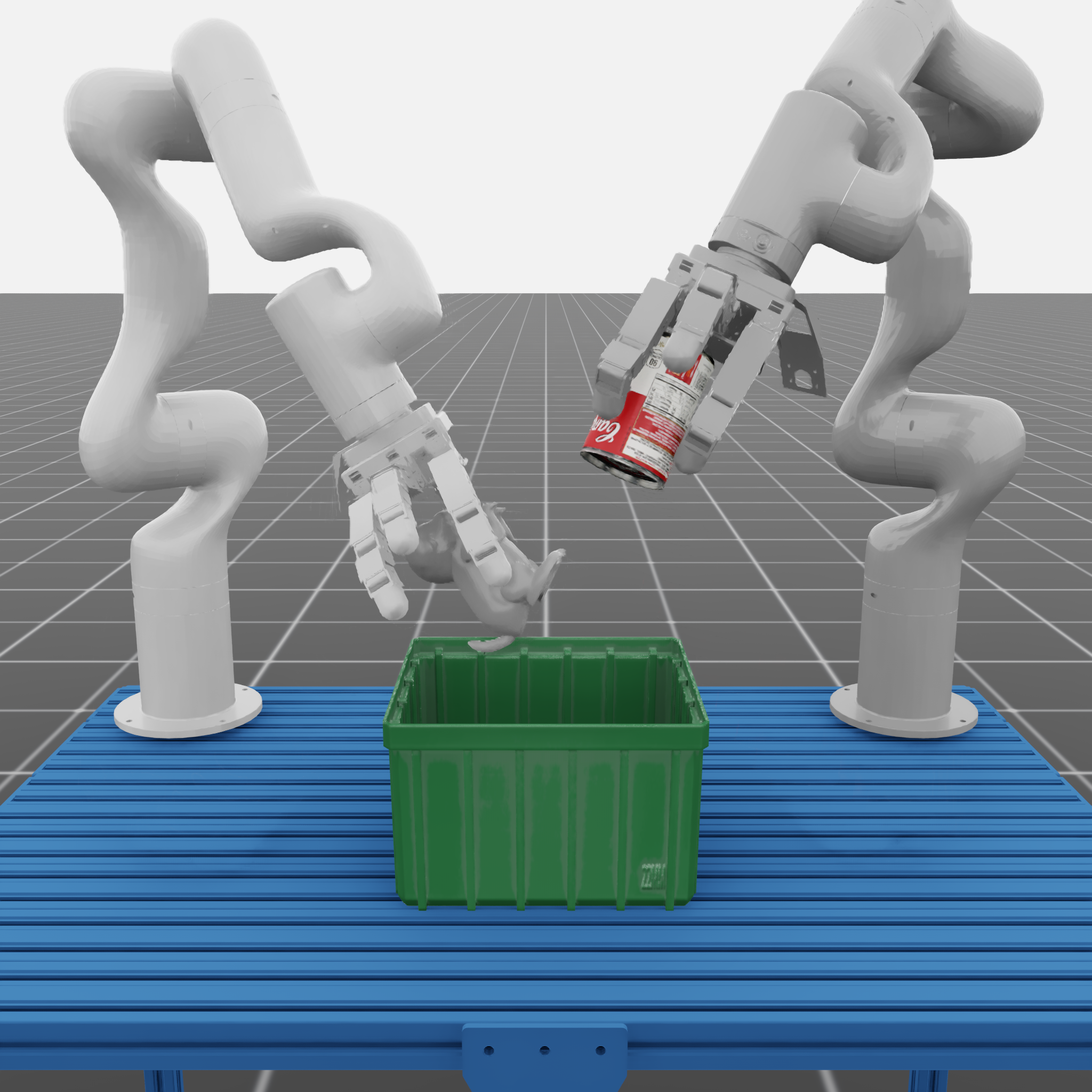}
\vspace{0.8em}
\end{minipage}
\begin{minipage}{0.7\textwidth}
\texttt{Table-clean}: The goal is to clean objects from the workbench by placing them into a basket. Subtask 1 involves directly picking and placing the object into the basket. Subtask 2 involves picking up the object, waiting until the other agent completes its task, and then placing the object. To avoid collisions, the hand closer to the basket is expected to place its object first, while the other waits until the first has finished. Thus, the hands must coordinate their timing. 
\vspace{0.8em}
\end{minipage}
\textbf{Success criteria}: Both objects must be successfully placed inside the basket without any collisions.\\
\textbf{Reward functions for both subtasks}: (1) A reaching reward between finger and the object; (2) An object distance reward to encourage moving the object toward the basket; (3) A success bonus for placing the object inside the basket. \\
\textbf{Additional reward for subtask 2}:
(4) A waiting reward to encourage proper timing and coordination with the other agent.
\\

\begin{minipage}{0.3\textwidth}
\centering
\includegraphics[width=0.65\linewidth]{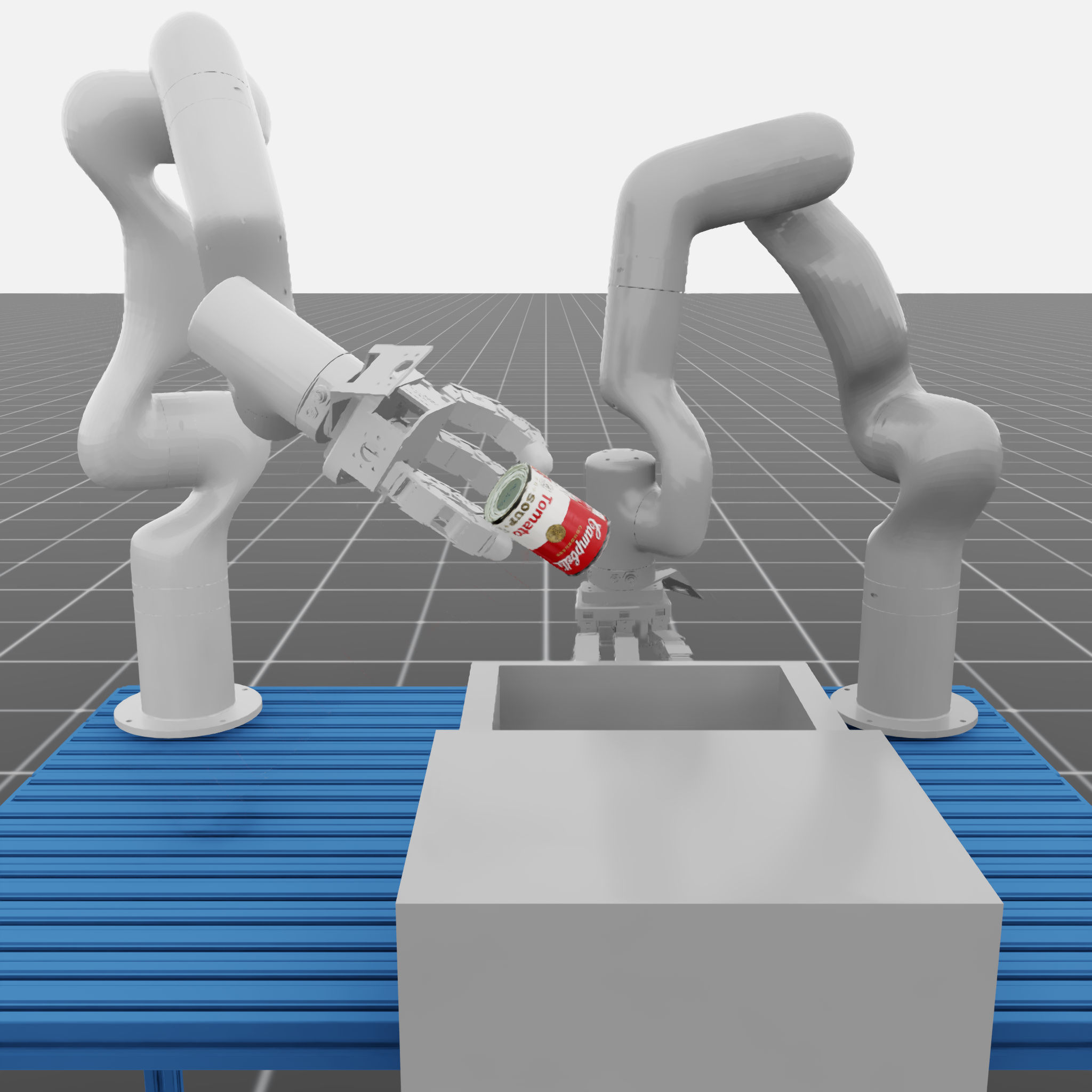}
\vspace{0.8em}
\end{minipage}
\begin{minipage}{0.7\textwidth}
\texttt{Drawer-insert}: The goal is to place an object into a drawer. Subtask 1 involves directly picking up the object and placing it into the open drawer. Subtask 2 involves pulling the drawer open, waiting until the object is placed inside, and then pushing the drawer closed. The subtasks are loosly coupled, therefore requiring minimal coordination.
\vspace{0.8em}
\end{minipage}
\textbf{Success criteria}: The object is inside the drawer, and the drawer is fully closed for $20$ consecutive steps.\\
\textbf{Reward functions for subtask 1}: (1) A reaching reward for between finger and the object; (2) An object distance reward to encourage moving the object toward the drawer; (3) A success bonus for placing the object inside the drawer. \\
\textbf{Reward functions for subtask 2}: (4) A pulling reward for opening the drawer; (5) A pushing reward for closing it.
\\

\begin{minipage}{0.3\textwidth}
\centering
\includegraphics[width=0.65\linewidth]{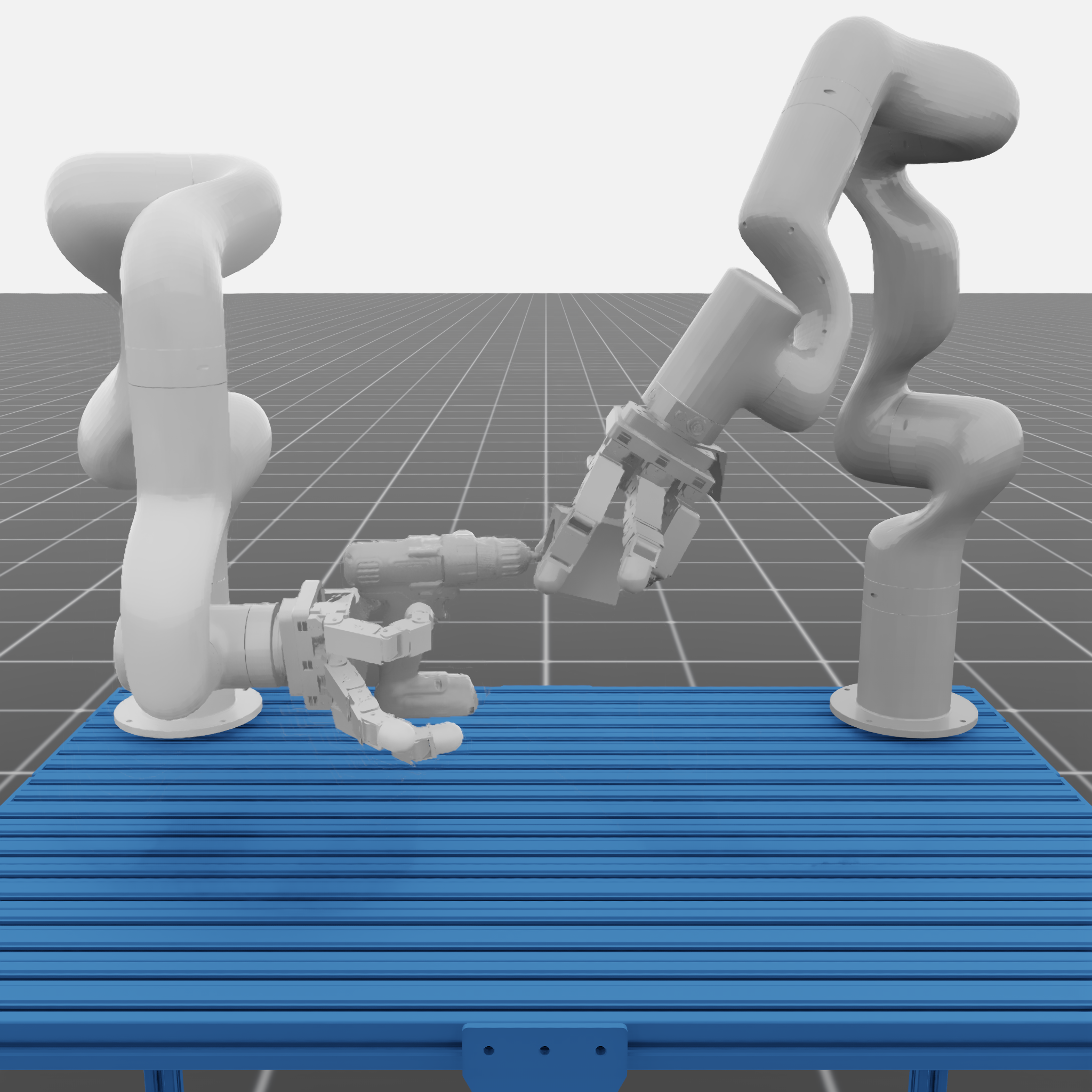}
\vspace{0.8em}
\end{minipage}
\begin{minipage}{0.7\textwidth}
\texttt{Threading}: The goal is to thread a drill into a holed cube in mid-air. Subtask 1 involves grasping the drill naturally and inserting its pin into the hole of the cube. Subtask 2 involves picking up the cube, reorienting it so that the hole faces the drill, and maintaining alignment.
This task requires precise bimanual coordination and synchronization for successful insertion.
\vspace{0.8em}
\end{minipage}
\textbf{Success criteria}: The drill pin must remain inside the cube’s hole for $20$ consecutive steps.\\
\textbf{Reward functions for subtask 1}:
(1) A hand alignment reward to align the palm with the drill;
(2) A drill pose reward to encourage lifting it to the correct mid-air position;
(3) A drill-cube distance reward to bring the drill closer to the cube.\\
\textbf{Reward functions for subtask 2}:
(4) A reaching reward to guide the fingers to the cube;
(5) A cube distance reward to move and align the cube with the drill.
\\

\begin{minipage}{0.3\textwidth}
\centering
\includegraphics[width=0.65\linewidth]{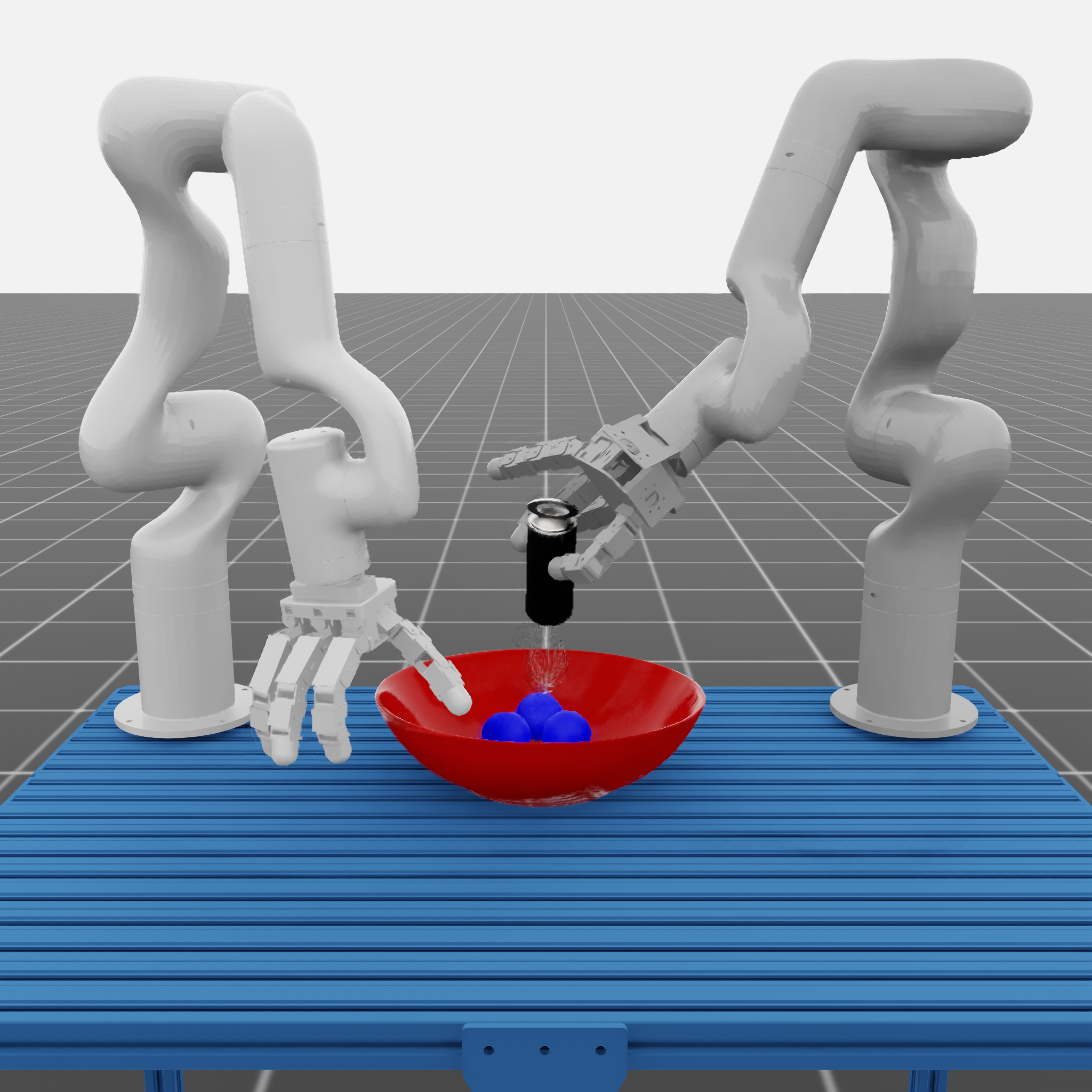}
\vspace{0.8em}
\end{minipage}
\begin{minipage}{0.7\textwidth}
\texttt{Bowl-stir}: The goal is to use an egg-beater to stir balls inside a bowl. Subtask 1 involves pushing the bowl to the center and stabilizing it for stirring. Subtask 2 involves picking up the egg-beater, reorienting it to face downward, and stirring the balls inside the bowl. This task emphasizes everyday dexterity, particularly the challenge of in-hand reorientation.
\vspace{0.8em}
\end{minipage}
\textbf{Success criteria}: The egg-beater must be aligned above the bowl and positioned correctly for stirring. \\
\textbf{Reward functions for subtask 1}:
(1) A hand alignment reward to align the palm with the bowl;
(2) A bowl pose reward to encourage centering and stabilization. \\
\textbf{Reward functions for subtask 2}:
(3) A reaching reward to guide the hand to the egg-beater;
(4) An egg-beater distance reward to position it correctly above the bowl;
(5) A stirring reward based on the motion (velocity) of the balls inside the bowl.
\\

\begin{minipage}{0.3\textwidth}
\centering
\includegraphics[width=0.65\linewidth]{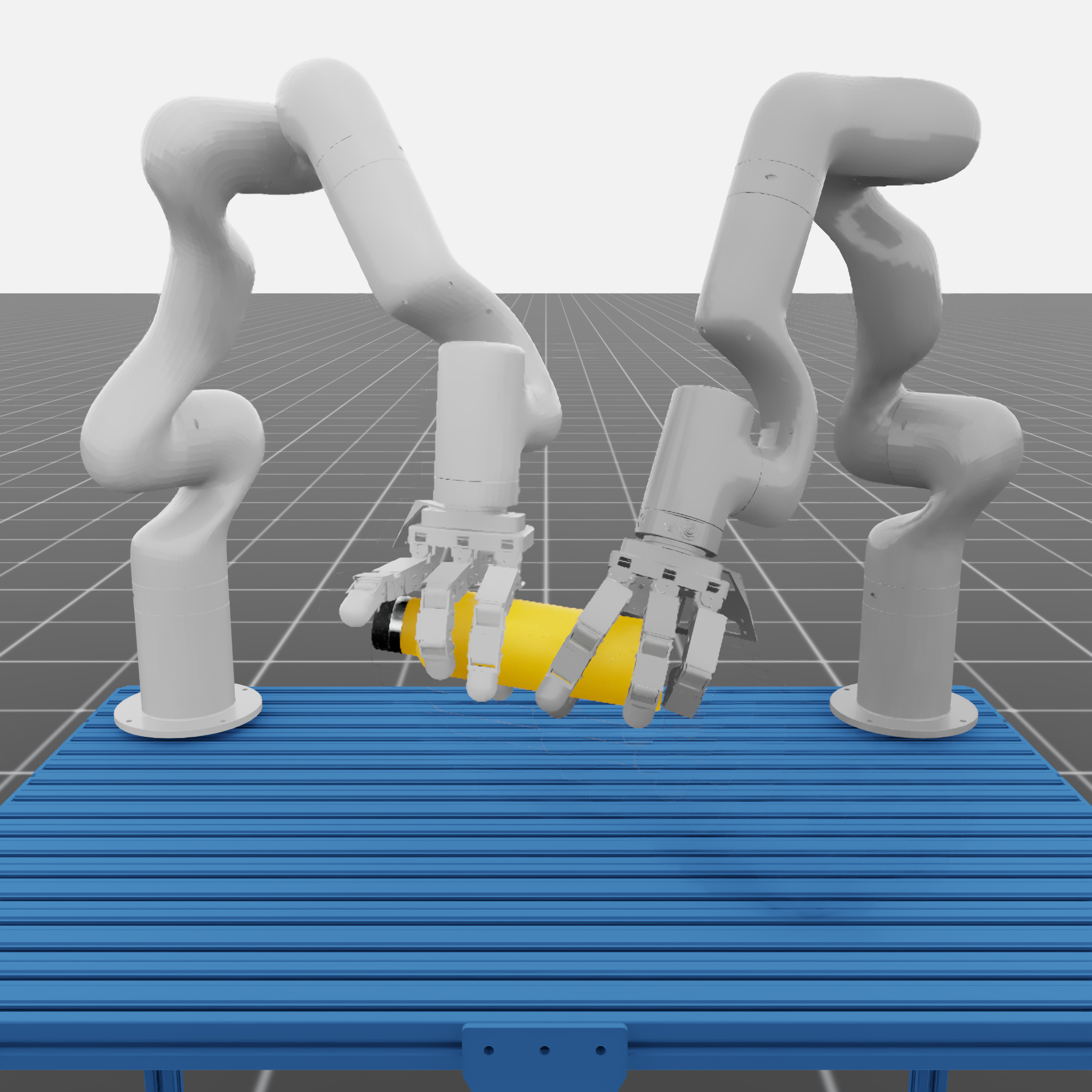}
\vspace{0.8em}
\end{minipage}
\begin{minipage}{0.7\textwidth}
\texttt{Handover}: The goal is to use the closer hand to grasp a bottle from the table and hand it over to the other hand. Subtask 1 involves grasping the bottle in a way that facilitates the handover and passing it to the other hand.
Subtask 2 involves receiving the bottle from the other hand and holding it steadily in mid-air. The main challenges lie in grasp planning and ensuring a smooth, coordinated transition between the hands.
\vspace{0.8em}
\end{minipage}
\textbf{Success criteria}: The hand farther from the bottle must hold it steadily for $20$ consecutive steps. \\
\textbf{Reward functions for subtask 1}:
(1) A reaching reward to guide the hand to the bottle;
(2) A bottle pose reward to encourage lifting it to the correct mid-air position;
(3) A releasing reward that penalizes excessive holding force, encouraging proper release during handover.\\
\textbf{Reward functions for subtask 2}:
(4) A hand alignment reward to align the palm with the bottle;
(5) A bottle pose reward to encourage stable holding after receiving the bottle.

\section{Real world System}
\label{app:real_world}
\paragraph{Robot Platform, Sensors and Control} Our robotic platform consists of two 6-\glspl{dof} xArm UF850 arms, each equipped with a 16-\glspl{dof} Allegro Hand V4, yielding a total of 44 degrees of freedom. The system operates over a 1.2$\times$0.8m tabletop workspace under standard safety constraints. A low-level joint-level PD controller runs at 120Hz, while policy inference is executed at 20Hz. Perception is provided by a single egocentric ZED2i RGB-D camera mounted between the arms. We integrate FoundationPose~\citep{wen2024foundationpose} and SAM2~\citep{ravi2024sam} for robust multi-object tracking.

\begin{figure}[t] \centering \includegraphics[width=1.\linewidth]{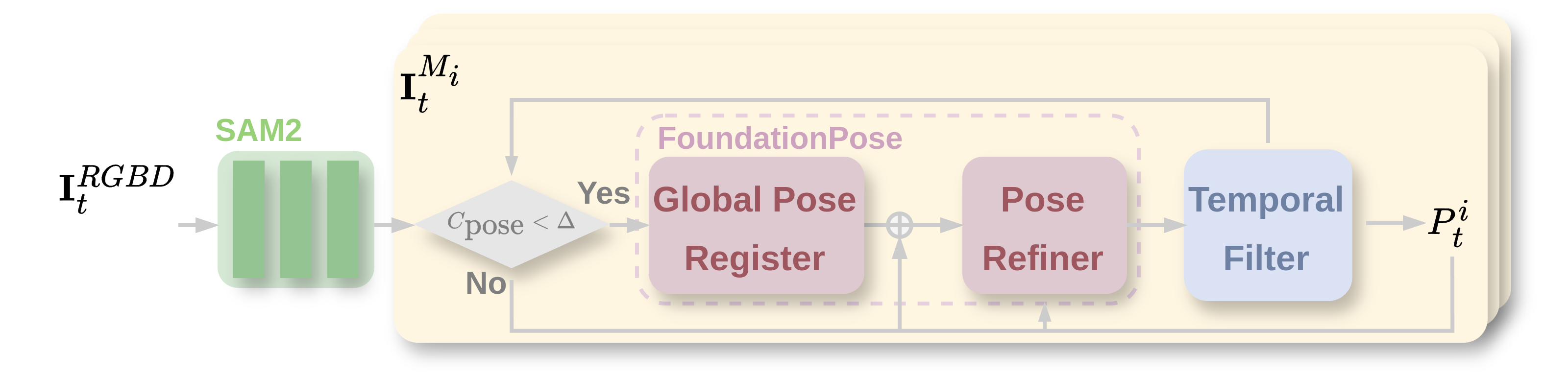} \caption{Overview of Perception} \label{fig:perception overview} \end{figure}

\paragraph{Perception Pipeline} Our perception pipeline (Fig.~\ref{fig:perception overview}) combines FoundationPose~\cite{wen2024foundationpose} and SAM2~\cite{ravi2024sam} to achieve robust, real-time 6D object pose tracking in cluttered and dynamic scenes. The input consists of RGB-D frames captured at 1080p and 30 FPS from a single ZED2i camera mounted between the robot arms. We operate the camera in ultra mode to maximize depth range and preserve Z-accuracy along the sensing axis, which is crucial for high-precision pose estimation.

For each incoming frame, FoundationPose is executed in parallel for all known objects to predict their 6D poses. While FoundationPose is robust under typical conditions and performs well on standard benchmarks, it fails to recover object pose when faced with rapid motion or complete occlusion.

To handle such cases, we integrate SAM2 for multi-object segmentation and tracking. For each object, we render expected RGB and depth images using a lightweight offscreen renderer based on the object’s CAD model. These rendered views are compared against the observed images from the ZED2i camera. Pose confidence is computed by measuring photometric and geometric discrepancies between the rendered and observed RGB-D images. Specifically, we define the confidence score as:
\begin{equation}
	c = \exp\left( -\frac{1}{\sum_{i} M^{(i)}} \sum_{i} M^{(i)} \left( \left\| I_{\text{obs}}^{(i)} - I_{\text{rend}}^{(i)} \right\|_1 + \lambda \cdot \left\| D_{\text{obs}}^{(i)} - D_{\text{rend}}^{(i)} \right\|_1 \right) \right)
\end{equation}

where $M^{(i)}$ is a binary foreground mask obtained from SAM2. $I_{\text{obs}}$, $I_{\text{rend}}$ denote observed and rendered RGB images, $D_{\text{obs}}$, $D_{\text{rend}}$ denote depth images, $N$ is the number of valid pixels, and $\lambda$ balances color and depth contributions. If the confidence $c < 0.5$, the object is deemed lost, and its pose is re-initialized using FoundationPose without relying on temporal priors.

To mitigate jitter and ensure smooth input to the policy, we apply SLERP interpolation~\cite{kremer2008quaternions} for rotations and linear interpolation for translations in SE(3) across consecutive pose estimates, followed by exponential moving average (EMA) filtering. This ensures temporally coherent trajectories and aligns the pose update rate with the policy’s inference frequency of 20 FPS.

\section{Baseline}
\label{app:baselines}
\begin{table}[htb!]
	\centering
	\small
	\begin{tabular}{lcccccc}
		\toprule
		Algorithms             & \acrshort{eppo} & \acrshort{ippo} & \acrshort{eippo} & \acrshort{smc} & \acrshort{smaug} & \textbf{SYMDEX (Ours)} \\
		\midrule
		\rowcolor[HTML]{ECF4FF}
		\# of Policies         & 1               & 2               & 1                & 2              & 2                & 2                      \\

		\# of Tasks per Policy & 2               & 2               & 2                & 1              & 1                & 1                      \\

		\rowcolor[HTML]{ECF4FF}
		Action Dim. per Policy & 44              & 22              & 22               & 22             & 22               & 22                     \\

		\# of Value Functions  & 1               & 2               & 1                & 1              & 2                & 2                      \\

		\rowcolor[HTML]{ECF4FF}
		Uses Equi. Network     & Yes             & No             & Yes              & Yes            & No               & Yes                    \\
		\bottomrule
	\end{tabular}
        \vspace{0.5em}
	\caption{Comparison design choices of the five baselines and SYMDEX.}
	\label{tab:baseline}
\end{table}

\section{Hyperparameters}\label{app:hyperparameters}
We list the hyperparameters used for all baselines. Since all methods, including \textsc{SYMDEX}, use \gls{ppo} as the backbone algorithm, they share identical hyperparameters to ensure a fair comparison. Additionally, we use the same set of hyperparameters across all tasks—except for the entropy coefficient (Tab.~\ref{tab:entropy})—highlighting the robustness of our method to hyperparameter tuning. As shown in the table, we generally recommend starting with an entropy coefficient of $0.01$ for new tasks. If the task does not require extensive exploration, this can be reduced to $0.005$.

\begin{table}[htb!]
\centering
\begin{tabular}{ll} 
\toprule
Hyperparameters                          & Values  \\ 
\midrule
Num. Environments                        & $4,096$   \\
Critic Learning Rate                     & $5 \times 10^{-4}$ \\
Actor Learning Rate                      & $3 \times 10^{-4}$ \\
Optimizer                                & Adam \\
Batch Size                               & $32,768$ \\
Horizon Length                           & $64$  \\
UTD Ratio                                & $8$    \\
Ratio Clip                               & $0.15$ \\
$\lambda$ for GAE                        & $0.95$ \\
Discount Factor ($\gamma$)               & $0.99 $ \\
Gradient Clipping                        & $0.5$  \\
Critic Network                           & $[256, 256, 256]$ \\
Actor Network                            & $[256, 256, 256]$ \\
\midrule
Curriculum: Threshold                    & $0.7$ \\
Curriculum: Update Freq.                 & $100$ \\
Curriculum: Total Step                 & $10$ \\
\bottomrule
\end{tabular}
\vspace{0.5em}
\caption{Hyperparameter setup for \textbf{all methods} and \textbf{all tasks}.}
\label{tab:hyperparameters}
\end{table}

\begin{table}[htb!]
\centering
\begin{tabular}{cc} 
\toprule
                     & Entropy Coefficient  \\ 
\midrule
\texttt{Box-lift}       & $0.0$        \\
\texttt{Table-clean}    & $0.005$         \\
\texttt{Drawer-insert}  & $0.01$         \\
\texttt{Threading}      & $0.01$        \\
\texttt{Bowl-stir}      & $0.01$        \\
\texttt{Handover}       & $0.005$         \\
\bottomrule
\end{tabular}
\vspace{0.5em}
\caption{The entropy coefficient used for six tasks.}
\label{tab:entropy}
\end{table}

We list the final domain randomization and safety penalty values in Tab.~\ref{tab:curriculum}. We split the curriculum into $10$ stages (as shown in Tab.~\ref{tab:hyperparameters}), where each stage increases the level of randomization and penalty scale, allowing the agent to adapt progressively. For every $100$ policy updates, we track the agent’s success rate during training; if the success rate exceeds a predefined threshold $0.7$, the agent advances to the next stage. By simplifying the environment in the early stages, the agent can first focus on mastering the core task before dealing with harder, more variable situations—enabling more stable and effective training.

\begin{table}[htb!]
\begin{adjustbox}{max width=\textwidth}
\begin{tabular}{lcccccc}
\toprule
\multicolumn{1}{c}{}     & \texttt{Box-lift}            & \texttt{Table-clean}         & \texttt{Drawer-insert}       & \texttt{Threading}       & \texttt{Bowl-stir}           & \texttt{Handover}      \\
\midrule
Obj. Mass(kg)            & $[0.1, 0.5]$ & $[0.02, 0.2]$ & $[0.02, 0.2]$ & $[0.1, 0.3]$ & $[0.02, 0.2]$ & $[0.1, 0.4]$ \\
Obj. Init. Pos.(cm)      & $+\mathcal{U}(-0.1, 0.1)$            &  $+\mathcal{U}(-0.1, 0.1)$             &  $+\mathcal{U}(-0.15, 0.15)$            &  $+\mathcal{U}(-0.05, 0.05)$           &  $+\mathcal{U}(-0.1, 0.1)$             &  $+\mathcal{U}(-0.1, 0.1)$        \\
Obj. Init. Orien.(rad.)  &  $+\mathcal{U}(-0.7, 0.7)$            &  $+\mathcal{U}(-1.57, 1.57)$            & $+\mathcal{U}(-1.57, 1.57)$            & $+\mathcal{U}(-0.75, 0.75)$          & $+\mathcal{U}(-0.6, 0.6)$          & $+\mathcal{U}(-1.57, 1.57)$          \\
Obj. Random Force(N)     & \multicolumn{6}{c}{$0.5$}                                                                                \\
\midrule
Static Friction          & \multicolumn{6}{c}{$[0.24, 1.6]$}                                                                    \\
Dynamic Friction         & \multicolumn{6}{c}{$[0.24, 1.6]$}                                                                    \\
Restitution              & \multicolumn{6}{c}{$[0.0, 1.0]$}                                                                     \\
\midrule
Obj. Pose Obs. Noise     & \multicolumn{6}{c}{$+\mathcal{U}(-0.01, 0.01)$}                                                                        \\
Joint Position Noise     & \multicolumn{6}{c}{$+\mathcal{N}(0, 0.005)$}                                                                     \\
\midrule
Energy Penalty Coeff.    & \multicolumn{6}{c}{$-0.001$}                                                                              \\
Collision Penalty Coeff. & \multicolumn{6}{c}{$-1000.0$}       \\                                    \bottomrule           
\end{tabular}
\end{adjustbox}
\vspace{0.5em}
\caption{Domain randomization and safety penalty setup.}
\label{tab:curriculum}
\end{table}

\section{Additional Experiments}
\label{app:additional_exp}
\subsection{Curriculum Learning}
\begin{figure}[h]
	\centering
	\includegraphics[width=0.92\textwidth]{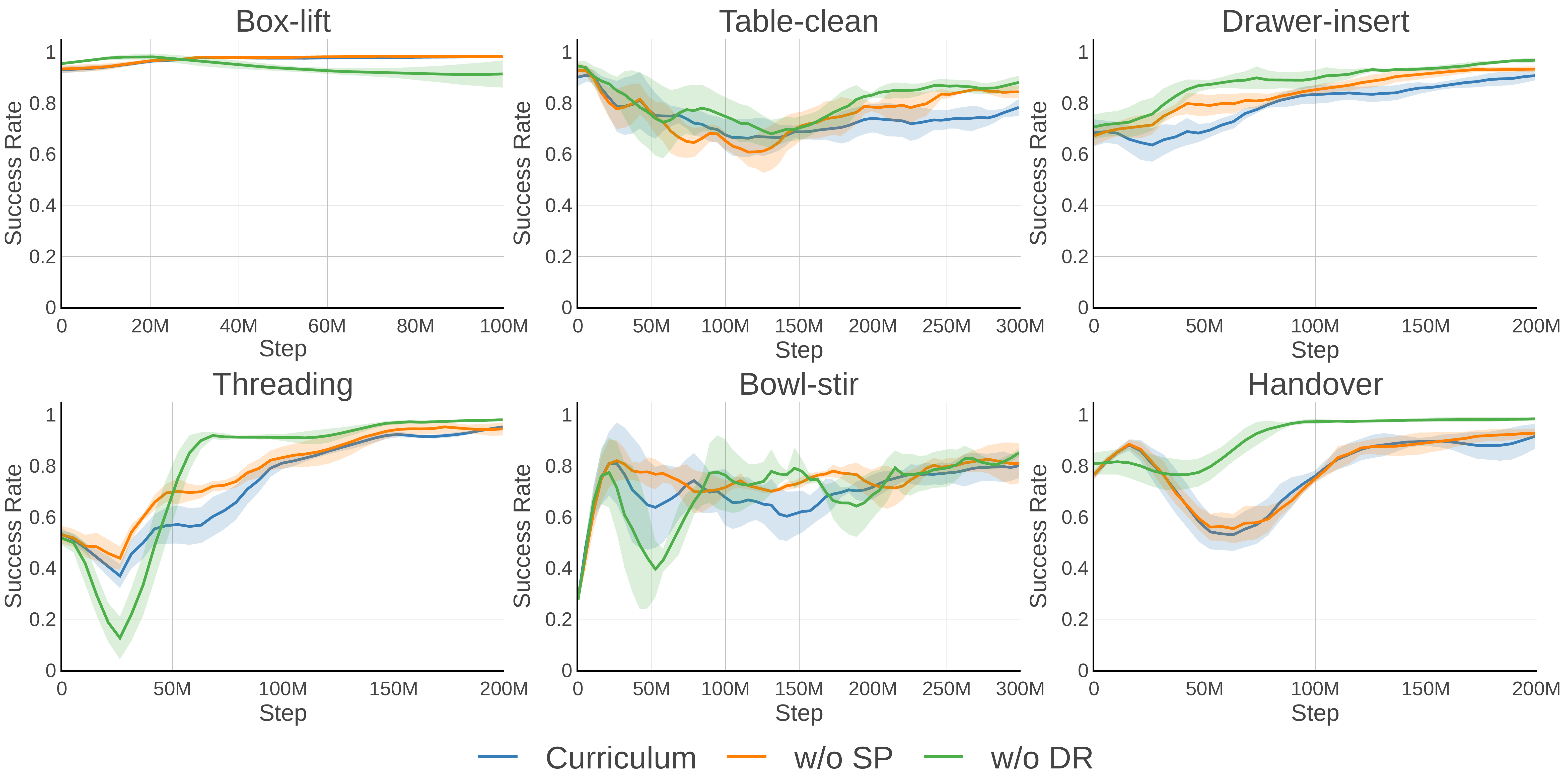}
	\vspace*{-1.5mm}
	\caption{
		Performance comparison of curriculum learning, curriculum w/o safety penalty (SP), and curriculum w/o domain randomization (DR) on six benchmark tasks.}
	\label{fig:performance_curriculum}
\end{figure}

We report the learning performance during the curriculum learning stage, which we treat as a separate phase. In this stage, we load the best checkpoint from initial training and perform fine-tuning. Since we use \gls{ppo} as the backbone algorithm, the fine-tuning process remains stable.

We evaluate the effectiveness of curriculum learning by comparing the full curriculum to ablations of its two key components: safety penalty and domain randomization. As shown in Fig.~\ref{fig:performance_curriculum}, the curriculum initially causes a performance drop across all six tasks, and then the performance is gradually improved as training progresses. We observe that the full curriculum converges more slowly, which is expected given it combines both components. Notably, the agent adapts more easily to the safety penalty than to domain randomization. In the \texttt{Box-lift} task, performance remains stable during the curriculum phase, since object randomization and collision penalties were already introduced in the initial training stage.

\subsection{Multi-arm Experiment}
\begin{figure}[h]
	\centering
	\includegraphics[width=\textwidth]{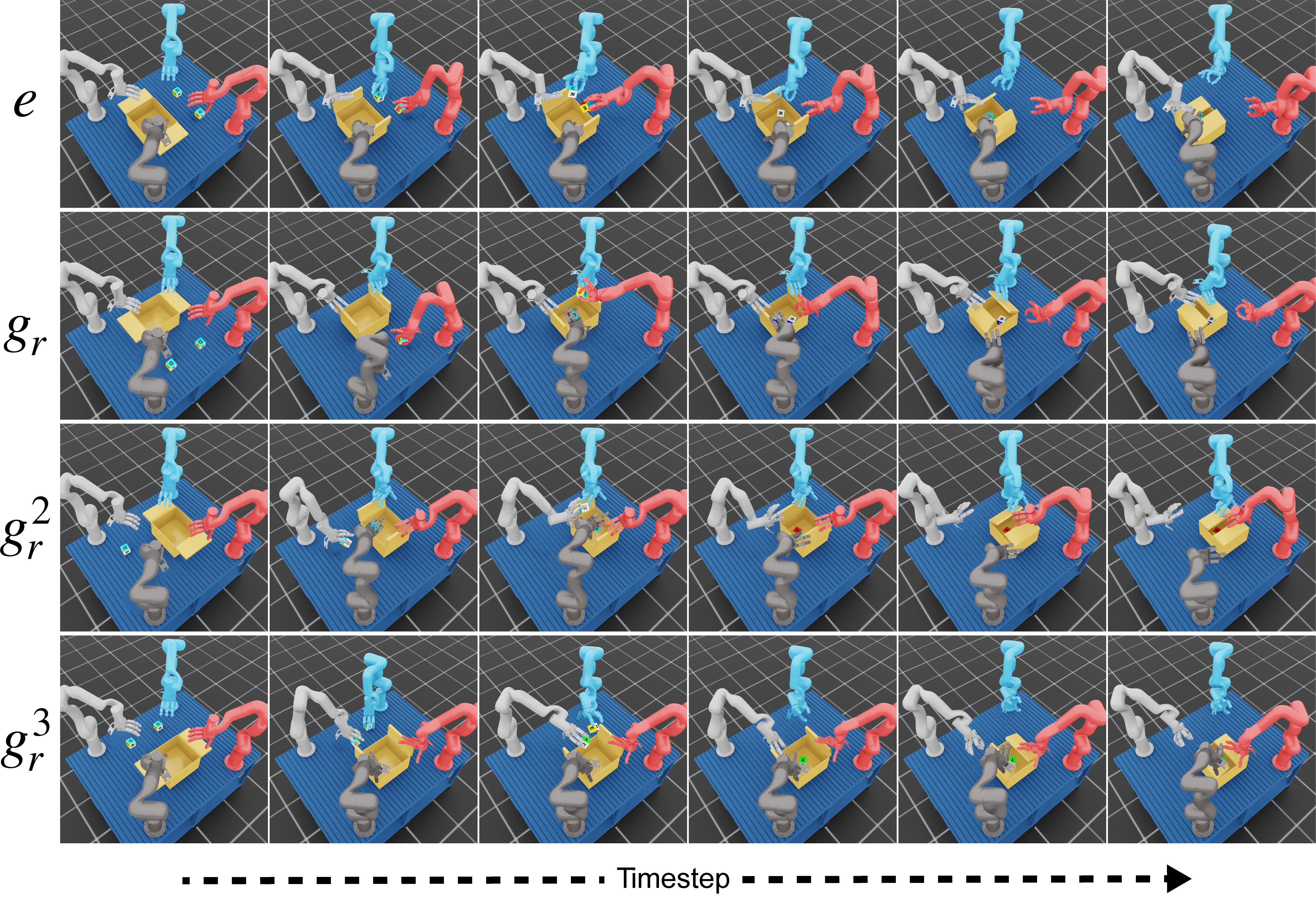}
	\vspace*{-1.5mm}
	\caption{Environment-policy rollout for the multi-arm task starting from state $\vs_0$ and all its symmetry states $g_r \Glact[\vsS] \vs_0$, $g_r^2 \Glact[\vsS] \vs_0$, and $g_r^3 \Glact[\vsS] \vs_0$. The four-arm system exhibits symmetry under the group $\G=\CyclicGroup[4]=\{e,g_r,g_r^2,g_r^3\mid g_r^4=e\}$, where $g_r$ is a $90^\circ$ rotation about the vertical axis.}
	\label{fig:4-arm}
\end{figure}

\begin{figure}[h]
	\centering
	\includegraphics[width=0.4\linewidth]{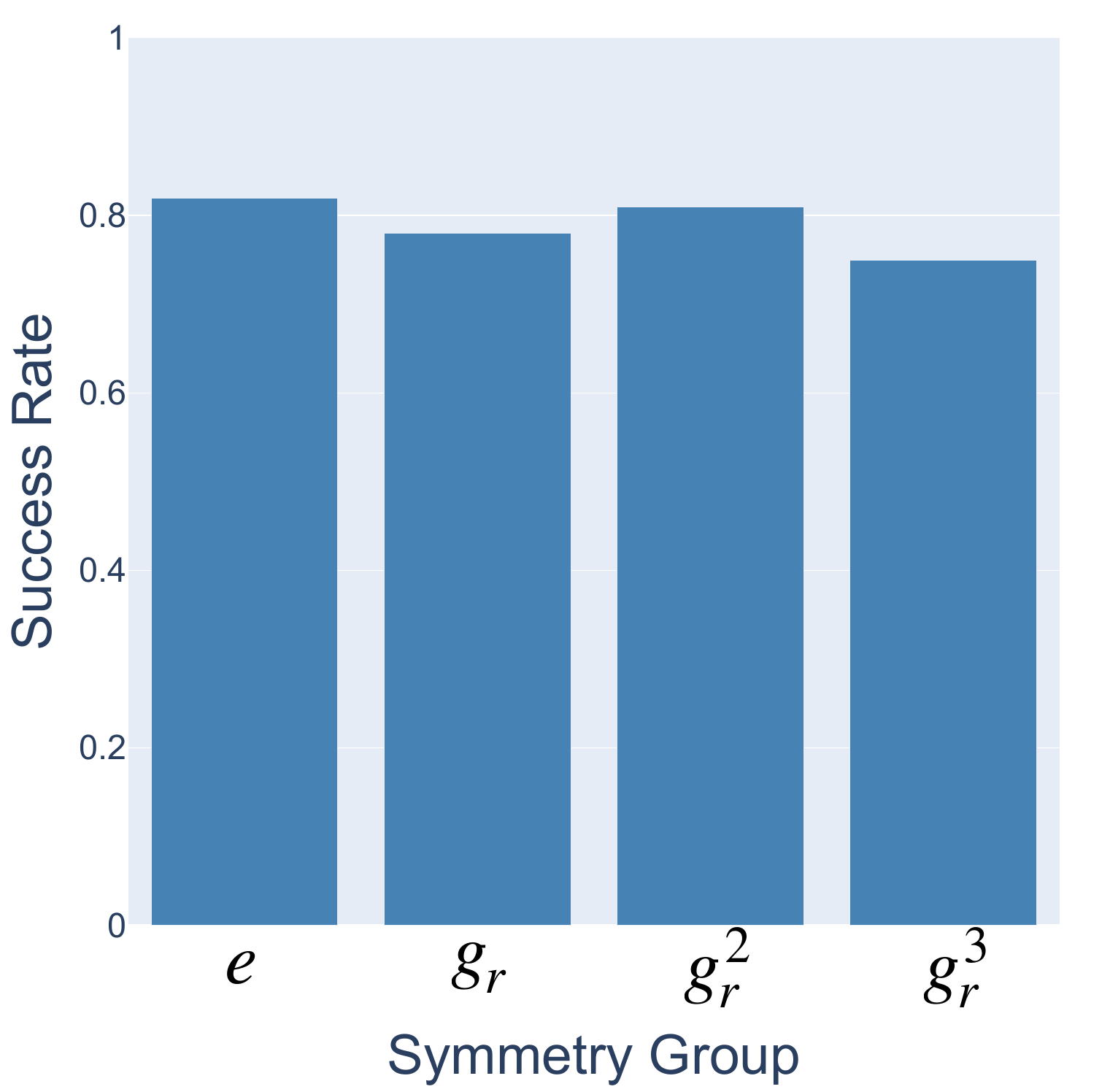}
    \vspace*{-1.5mm}
	\caption{Performance of \textsc{SYMDEX} on the multi-arm task.}
	\label{fig:4_arm_success_rate}
	\vspace*{-0.4cm}
\end{figure}

We visualize the environment-policy rollouts across all symmetry groups defined by $\G=\CyclicGroup[4]=\{e,g_r,g_r^2,g_r^3\mid g_r^4=e\}$, as shown in Fig.~\ref{fig:4-arm}. The first column shows the initial states, where the object configurations are rotated by $90^\circ$ about the vertical axis across different symmetry groups. As the policy executes, we observe symmetric behaviors generated by the equivariant policy. Although the four colored arms remain fixed, \textsc{SYMDEX} successfully solves all configurations with consistent performance, as shown in Fig.~\ref{fig:4_arm_success_rate}.

\newpage
\subsection{Real World Experiment}
\begin{figure}[t]
	\centering
	\includegraphics[width=\textwidth]{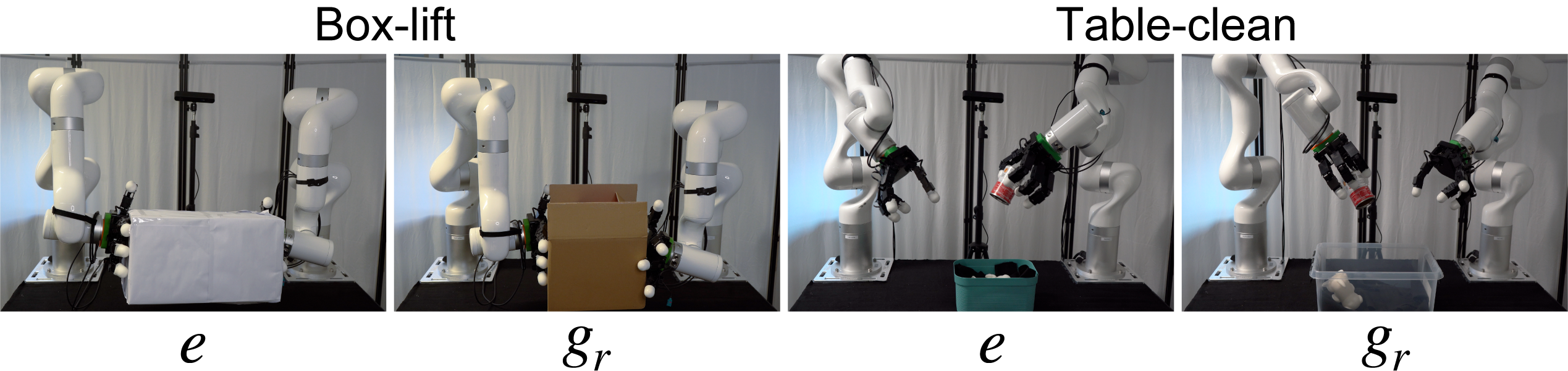}
	\vspace*{-1.5mm}
	\caption{Snapshots from the real-world experiments.}
	\label{fig:real-world-figure}
\end{figure}
We provide snapshots from real-world experiments on \texttt{Box-lift} and \texttt{Table-clean}, as shown in Fig.~\ref{fig:real-world-figure}, covering both original and symmetric scenarios. In \texttt{Box-lift}, both agents manipulate the same object and perform identical subtasks, so there is no significant difference between the original and symmetric settings.

Additionally, we evaluate our policy on out-of-distribution (OOD) objects. For example, in \texttt{Box-lift}, we use boxes of varying sizes that were never seen during training; in \texttt{Table-clean}, we introduce an OOD toy dog. Thanks to the curriculum learning strategy, our policy generalizes well and successfully handles these OOD cases.

\setacronymstyle{long-short}
\printglossary[type=\acronymtype, title={Acronyms}, toctitle={Acronyms}]

\end{document}